# CANDLE: A Cross-Modal Agentic Knowledge Distillation Framework for Interpretable Sarcopenia Diagnosis

Yuqi Jin[1†], Zhenhao Shuai[2†], Zihan Hu[1], Weiteng Zhang[3], Hai Lin[2], Weihao Xie[1],

Jianwei Shuai[2*], Xian Shen[3*], Zhen Feng[1,2*]

*1 College of Information and Engineering, First Affiliated Hospital,*

*Wenzhou Medical University, Wenzhou, China*

*2 Wenzhou Institute, University of Chinese Academy of Sciences, Wenzhou, China*

*3 Department of Gastrointestinal Surgery, First Affiliated Hospital,*

*Wenzhou Medical University, Wenzhou, China*

*† These authors contributed equally to this work*

**Correspondence:*

*shuaijw@wiucas.ac.cn (J.-W. S.); shenxian@wmu.edu.cn (X.S.); zfeng2019@foxmail.com (Z.F.)*

**Abstract— Background and Aims: Large language models (LLMs) have shown remarkable generalization and transfer capabilities by learning from vast corpora of text and web data. Their semantic representations allow cross-task knowledge transfer and reasoning, offering promising opportunities for data-scarce and heterogeneous domains such as clinical medicine. Yet, in diagnostic tasks like sarcopenia, major challenges remain: interpretability, transparency, and deployment efficiency. Traditional machine learning (TML) models provide stable performance and feature-level attribution, ensuring traceable and auditable decision logic, but lack semantic breadth. Conversely, LLMs enable flexible inference but often function as opaque predictors. Existing integration strategies remain shallow, rarely embedding the structured reasoning of TML into LLM inference.**

**Methods: Using sarcopenia diagnosis as a case study, SHapley Additive exPlanations (SHAP) were extracted from a baseline XGBoost model and transformed into structured, LLM-compatible representations. An actor-critic reinforcement learning (RL) strategy guided the LLM to reason over these SHAP-based inputs, producing calibrated rationales and refined decision rules. The distilled reasoning was consolidated into a structured knowledge repository and deployed via retrieval-augmented generation (RAG) for case-based inference.**

**Results: CANDLE achieved an accuracy of 79.3, exceeding the baseline XGBoost by 6 percentage points while preserving its confusion pattern alignment. Consistency analysis demonstrated high reproducibility: intraclass correlation coefficient (ICC) = 0.956 for probability outputs, Cohen's kappa = 0.842 for binary classifications, and a sample-level consistency rate of 95.5%. Across repeated runs under fixed prompts, the model maintained stable discrimination (AUC = 0.760 ± 0.010) and balanced sensitivity–specificity, with an optimal threshold of 0.697 ± 0.037. Prompt-specific evaluations revealed that while AUC remained stable, false positive and true positive rates varied with prompt framing, indicating that the model dynamically adapts its decision boundary rather than merely replicating XGBoost outputs.**

**Conclusion: By coupling SHAP-derived statistical evidence with reinforcement-trained LLM reasoning, CANDLE mitigates the interpretability–performance trade-off, enhances predictive accuracy, and preserves high decision consistency. The framework offers a scalable approach to knowledge assetization of TML models, enabling interpretable, reproducible, and clinically aligned decision support in sarcopenia and potentially broader medical domains.**

# I. Introduction

With the rapid development of large language models (LLMs), these systems have demonstrated strong generalization and transfer capabilities by learning from vast corpora of text and web data. Their semantic representations enable cross-task knowledge transfer and reasoning, offering new opportunities for data-scarce and heterogeneous domains. However, in clinical applications, challenges remain regarding interpretability, transparency, and deployment efficiency.

For decades, traditional machine learning (TML) methods, such as linear regression, decision trees, support vector machines, and ensemble approaches, have achieved stable and verifiable performance on both structured and partially unstructured data. Representative examples include SIFT for object recognition [1], hidden Markov models for continuous speech decoding [2], phrase-based statistical machine translation [3], support vector machines for breast cancer detection [4], and ensemble methods for credit risk and asset return prediction [5]. In clinical research and deployment, TML models are often preferred due to their relative transparency, ease of auditing, and regulatory compliance.

When equipped with modern interpretability toolkits, TML models can extend their well-established robustness into greater clinical impact—pairing predictive performance with enhanced transparency, auditability, and regulatory alignment. Post-hoc frameworks such as SHapley Additive exPlanations (SHAP) provide both local and global feature attributions [6], strengthening the interpretability that underpins reproducibility and clinician trust [7-8]. In parallel, LLMs and other foundation models offer a complementary strength: the ability to encode broad domain knowledge in natural language or structured representations, enriching downstream classifiers with semantic context [9]. This complementarity has motivated recent work on hybrid pipelines that combine TML and LLMs—either by invoking external models as decision modules or collocating LLMs alongside domain-specific classifiers [10-14].

However, most current integrations remain superficial. They often pass TML outputs to LLMs as static inputs, without embedding the interpretable decision logic of TML systems into the LLMs' reasoning process. This shallow coupling limits deep, bidirectional knowledge transfer between interpretable models and generative reasoning frameworks, leaving much of the potential synergy untapped.

Bridging the decision logic of TML with the semantic reasoning of LLMs presents a promising direction for building clinically deployable, interpretable models. However, systematic approaches for transforming TML decision evidence (e.g., SHAP-based feature attributions) into structured knowledge that LLMs can actively internalize—and subsequently distill into smaller, resource-efficient models—remain scarce. Existing methods largely rely on output-level fusion or loose ensemble, without offering explicit and optimizable training strategies that preserve both interpretability and discriminative power.

To address this gap, we propose **CANDLE (C**ross-modal **A**gentic k**N**owledge **D**istillation framework for **L**LM-**E**nhanced diagnosis**)**, a framework designed for interpretable clinical diagnosis. CANDLE transforms the decision logic of TML models into structured, language-aligned evidence units (e.g., SHAP-based attributions and probabilistic outputs) and integrates them into the reasoning chain of LLMs. An Actor-Critic reinforcement learning (RL) calibration loop enables the LLM to actively absorb and internalize this evidence, which is then selectively distilled into compact diagnostic models. Applied to sarcopenia diagnosis as a representative cross-modal clinical task, CANDLE unifies imaging features and clinical measurements within an interpretable, semantically aligned representation—yielding models that balance diagnostic performance, medical transparency, and deployment feasibility.

# II. Related work

Recent research has increasingly explored the integration of machine learning (ML) and large language models (LLMs), aiming to combine the statistical modeling power of ML with the semantic reasoning and generative capabilities of LLMs. A common approach treats LLMs as semantic encoders and ML as classifiers. Huang et al. [15] employed LLM-based text embeddings with ML classifiers to predict life satisfaction, while Hong and Oh [16] used LLMs to extract semantic and structural features and applied ML for feature optimization in threat classification tasks.

Another line of work embeds ML or deep learning models as tool modules within LLM-driven pipelines. Wu et al. [17] incorporated ResNet and ViT as vision encoders into LLaMA for multimodal generation, and Manjunatha and Mahendra [18] combined CNN-based local feature extraction with transformer-based reasoning for tumor analysis. Similarly, Luo et al. [19] proposed an automated multimodal ML framework in which LLMs dynamically orchestrate heterogeneous ML models for different data modalities.

LLMs have also been leveraged for automated feature engineering and pipeline construction. Hollmann et al. [20] introduced CAAFE, where LLMs generate feature engineering code that is validated by ML models. Luo et al. [19] extended this idea by using LLMs to build dynamic AutoML pipelines across multimodal datasets. Beyond prediction, the synergy between ML and LLMs has been explored for interpretability. Khediri et al. [21] combined SHAP-based attributions with LLM-generated explanations for intrusion detection, while Oh et al. [22] applied SHAP and CatBoost to perioperative stroke prediction with natural language risk descriptions.

However, most of these efforts remain loosely coupled: either LLMs provide semantic representations for ML, or ML produces feature importance later translated by LLMs. Recent attempts such as verbalized ML [23] show a trend toward tighter integration by mapping ML parameter spaces into natural language, but they do not incorporate attribution-level reasoning. In contrast, our work directly injects SHAP-based attribution knowledge from ML models into LLM decision-making. This design bridges statistical feature attribution with semantic reasoning, enabling LLMs not only to verbalize but also to reason upon interpretable evidence derived from ML.

# III. Methods

## *2.1 Data processing*

We investigated the China Health and Retirement Longitudinal Study (CHARLS), a nationwide survey examining health and retirement trends among middle-aged and elderly individuals in China [24]. To achieve model balance, we select 300 records from normal individuals randomly and combine them with all 300 records of sarcopenia patients. Additionally, to guarantee the effectiveness of the model and training efficiency, we choose features based on feature importance. These features consist of 11 physical characteristics and 4 mental characteristics. For further specific dataset details, please refer to Appendix A.0.

### *2.2 Overall Framework of CANDLE*

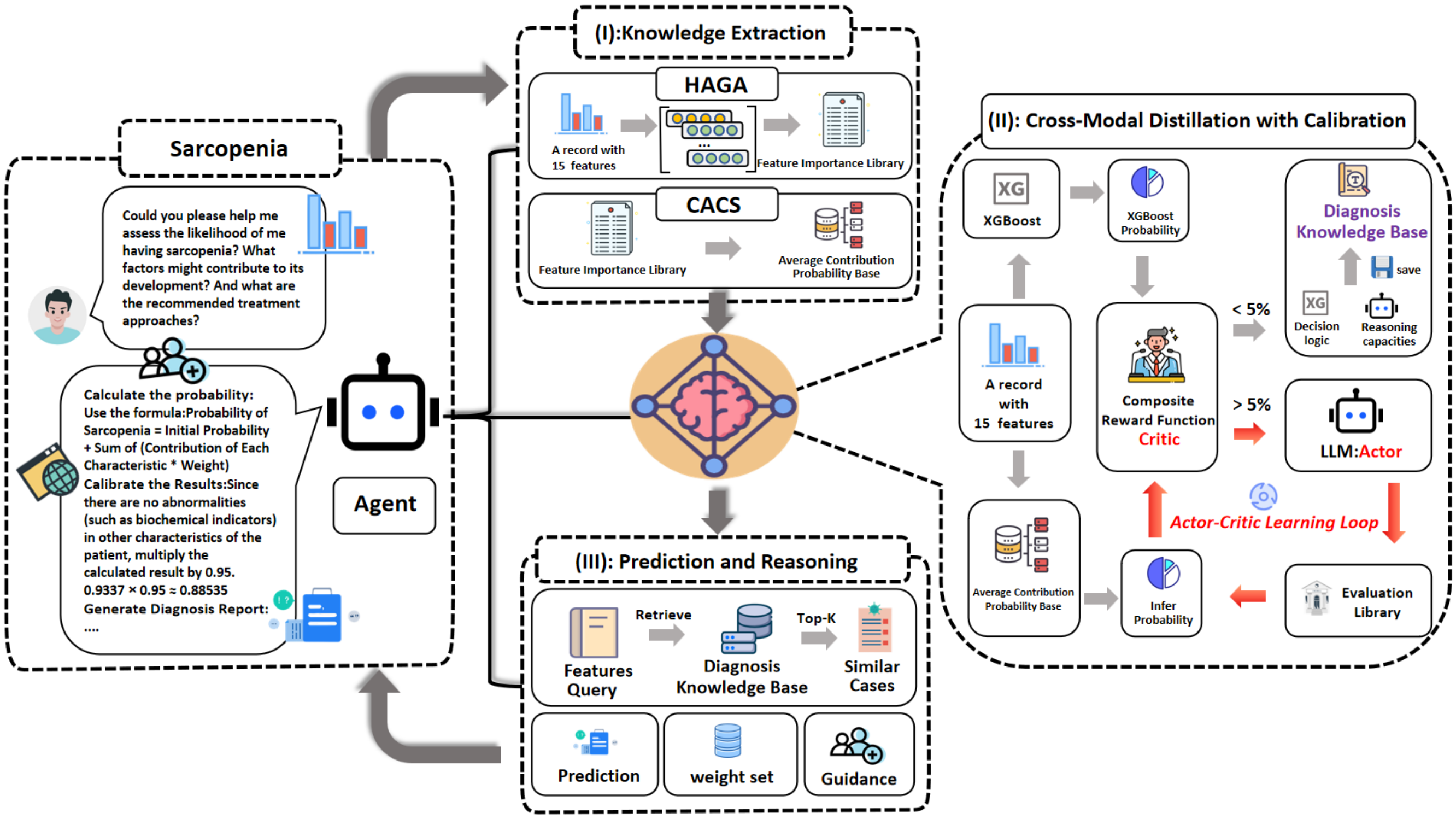

**Figure 1. The Framework of the CANDLE System.** The system is composed of three core modules: (I) Knowledge Extraction, which trains an XGBoost model and extracts its feature knowledge into an Average Contribution Probability Base (ACPB); (II) Cross-Modal Distillation, which uses an Actor-Critic reinforcement-learning loop to align Large Language Model's (LLM) outputs with the XGBoost model's predictions, generating a Diagnosis Knowledge Base (DKB); and (III) Prediction and Reasoning, where for a new case, similar precedents are retrieved from the DKB to help the LLM generate a final prediction and actionable guidance.

Figure 1 shows the three stages of the framework:

**Module I- Knowledge Extraction:** The knowledge extraction process begins with training an XGBoost model on the original dataset, from which only samples with correct predictions are retained. Using an explainability tool (e.g., SHAP), a "Feature Importance Library" is constructed to quantify each feature's predictive contribution. These importance metrics are then transformed into contribution probabilities, which are aggregated to form an Average Contribution Probability Base (ACPB)—a structured representation optimized for interpretability, specifically to enable utilization by LLMs in subsequent stages. This transformation, while introducing information degradation through reduced individual feature distinctiveness, intentionally creates latent space to facilitate the integration of inherent LLM knowledge into the decision-making process.

**Module II- Cross-Modal Distillation with Calibration:** Building on the latent space from Module I, this module introduces a Cross-Modal Distillation with Calibration framework to integrate LLMs with XGBoost. First, XGBoost's output probabilities are compared with the inferred probabilities from the ACPB using initial weights, setting up an Actor-Critic mechanism to align both models in the latent space.

In each iteration, the Actor updates the weights applied to the feature-level contribution probabilities in ACPB, while the Critic measures how close the reweighted probabilities are to XGBoost's outputs. This feedback loop gradually adjusts the weights so that the feature contributions in the latent space better match XGBoost's probabilistic patterns.

Meanwhile, the LLM incorporates its semantic knowledge into this calibration process. The final weights thus capture both the LLM's reasoning and XGBoost's statistical insights, serving as numerical regulators of feature contributions for downstream diagnostic tasks.

**Module III- Prediction and Reasoning:** For a newly arrived case, the system first retrieves the most similar records from the Diagnosis Knowledge Base (DKB) based on feature values. These retrieved cases containing feature values and optimized weights, together with the new case's feature values, are incorporated into a carefully designed prompt and fed into the LLM. Guided by this prompt and the precedents, the LLM infers a set of context-specific feature weights it deems most appropriate. These weights are then applied to the ACPB to compute the sarcopenia probability for the case, alongside generating a class prediction and actionable clinical guidance.

### *2.3* Module I: Knowledge Extraction

Within Module I focusing on Data & Machine Learning Knowledge Extraction, the construction of the ACPB is a core step that bridges model interpretability and LLM utilization. Building on the XGBoost model trained on the original dataset—with only correctly predicted samples retained, 324 in this dataset—and the "Feature Importance Library" derived using SHAP to quantify each feature's predictive contribution and create latent space for LLMs further using, the process proceeds as follows:

#### *2.3.1 Half-step Aligned Group Averaging (HAGA)*

Feature Grouping by Intervals**:** The Feature-SHAP Matrix undergoes transposition to generate distinct feature sets, each encompassing 324 feature values paired with their corresponding SHAP values. Feature values are then discretized into non-overlapping intervals via a unified partitioning rule:

For continuous features, equidistant midpoints (step size = 0.5) define closed intervals of width 0.5 (e.g., midpoint 0.5 corresponds to [0.25, 0.75]), ensuring all values within ±0.25 of the midpoint are included.

For integer features, intervals are defined by exact matching with integer midpoints, such that a feature value is assigned to a group only if it strictly equals the midpoint.

This discretization ensures bijective mapping between feature values and intervals, transforming the raw feature space into a structured set of discrete groups.

Interval Mean Aggregation**:** To resolve ambiguity in associating new feature values with interval groups (wherein multiple values may fall within a single interval, introducing ambiguity in matching), the arithmetic mean of SHAP values within each interval is computed as the representative metric. This process yields feature-specific SHAP knowledge sub-bases (one per feature), each containing discretized feature ranges and their corresponding average SHAP values, thereby aligning individual instances with subgroup-level SHAP distribution patterns.

#### 2.3.2 Probabilistic Transformation for ACPB

To convert SHAP-based importance metrics into a format optimized for LLM utilization, a probabilistic transformation framework is implemented, consisting of the following steps: The probabilistic transformation framework first normalizes raw SHAP values to the [0, 1] range using the sigmoid function—converting feature contributions into interpretable probabilistic measures that are more accessible for LLM utilization. Building on this, the Contrastive Attribution Calculation via Sigmoid (CACS) leverages the model's base value to compute the differential probability contribution of each interval's average SHAP value. This two-step process collectively transforms SHAP-based feature importance into interval-specific contribution probabilities, aligning the quantitative model outputs with a structured format optimized for LLM understanding and subsequent knowledge integration. Each probability is defined as an atomic fact, meaning that a given feature value uniquely corresponds to a single probability contribution, forming the minimal evidence unit directly usable by LLMs without additional inference.

#### 2.3.3 ACPB Construction

Aggregation of these contribution probabilities across all intervals results in the ACPB. Structured as feature-specific sub-bases (one per feature, each containing interval-specific $p_j$ values), the ACPB balances interpretability and information compression—intentionally reducing individual feature distinctiveness to create latent space for integrating LLM-inherent knowledge in subsequent decision-making stages.

This structured transformation enables the ACPB to serve as a bridge between model-derived feature contributions and LLM capabilities, facilitating evidence-based sarcopenia assessment. For further specific details, please refer to Appendix A.1.

### 2.4 Module II: Cross-Modal Distillation

Module (II) is dedicated to cross-model knowledge distillation, with its core objective being the integration of XGBoost's decision-making logic and the reasoning capabilities of LLMs for the generation of diagnostic texts, achieved through the adoption of a continuous Actor-Critic loop. The workflow is as follows:

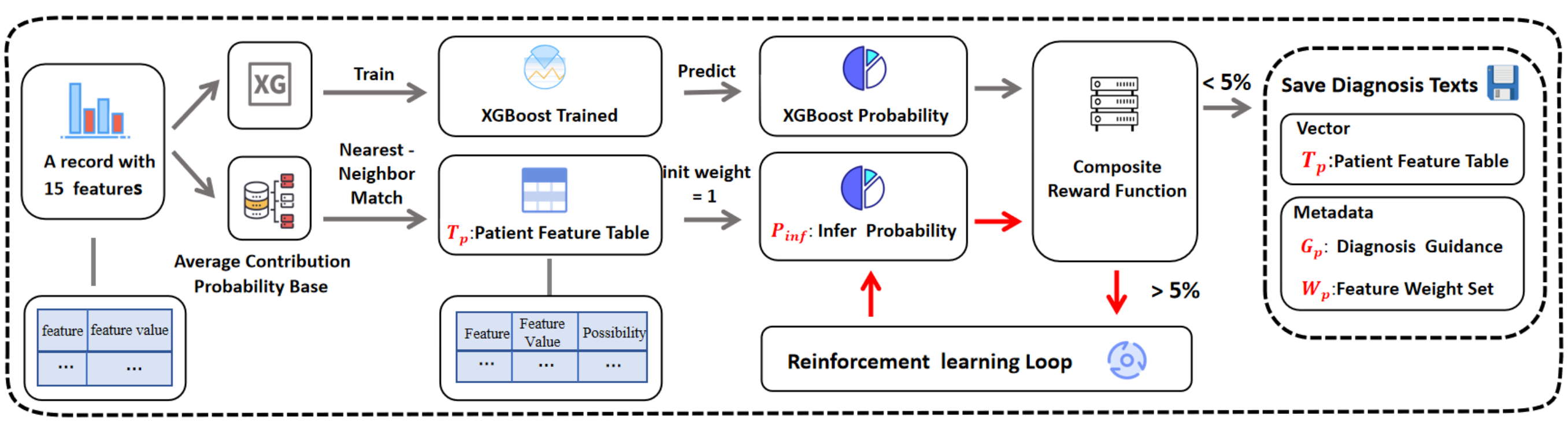

**Figure 2. Illustration of Actor-Critic Reinforcement Learning Loop for knowledge distillation**, detailing Module (II) of the framework. For a given patient case, an initial Infer Probability is computed using the ACPB and an initial weight set. This value is then evaluated against the XGBoost Probability by the Composite Reward Function (serving as the Critic). If their discrepancy exceeds 5%, the RL loop is activated and will continue running until the discrepancy falls below 5%

The proposed framework initiates with the extraction of structured clinical features from electronic patient records. These features are mapped to the Average Contribution Probability Base (ACPB) to obtain feature-level contribution probabilities, which together form a Patient Feature Table serving as the foundation for probabilistic inference in the latent space established in Module I.

A Composite Reward Function, acting as the Critic, assumes two principal responsibilities: Discrepancy Quantification: measuring the divergence between the reweighted inferred probabilities and XGBoost-generated probabilities; Reward Library Construction: compiling failure-case replay, probability-distance metrics, and directional guidance signals to guide the optimization process.

During the initial computation cycle, all weights applied to ACPB feature contributions are set to unity, representing the absence of any LLM-induced modulation. If the initial discrepancy falls below a 5% threshold, the system bypasses iterative refinement and directly archives both the weights and associated metadata into the DKB.

When the discrepancy exceeds the threshold, the Actor–Critic calibration loop is triggered. The Critic provides targeted feedback signals from the Reward Library; combined with feature-level descriptions and prompts, these signals guide the Actor (the LLM) to update the weights applied to ACPB's feature-level contribution probabilities. These weights serve exclusively as numerical regulators in the latent space, iteratively refined until the inferred probabilities achieve alignment with XGBoost's probabilistic outputs.

Upon convergence, only the optimized weight configurations—reflecting both XGBoost's statistical rigor and the LLM's semantic reasoning—are retained in the latent space for downstream tasks, with its metadata for retrieve. Detailed specifications for feature extraction, probabilistic inference, reward construction, and iterative optimization are presented in Appendix A.2, with pseudocode provided in Appendix A.5 (Algorithm 1).

### 2.5 Module III: Prediction and Reasoning

This method presents an integrated diagnostic framework combining. Feature-Grouped Multi-Round Retrieval for Intersection Case Selection and LLM-Driven Joint Analysis and Weight Adjustment, shown in Figure 4. First, clinical features are divided into three subsets for parallel case retrieval, with intersecting results ensuring cross-feature consistency. The system then employs LLMs to: (1) generate initial diagnoses, (2) refine them through similar case comparisons, and (3) produce adaptive feature weights. (4) Finally, calculate the sarcopenia probability and generate the diagnosis report. This approach enables precise, interpretable diagnoses by bridging statistical patterns with clinical reasoning, shown in figure 3

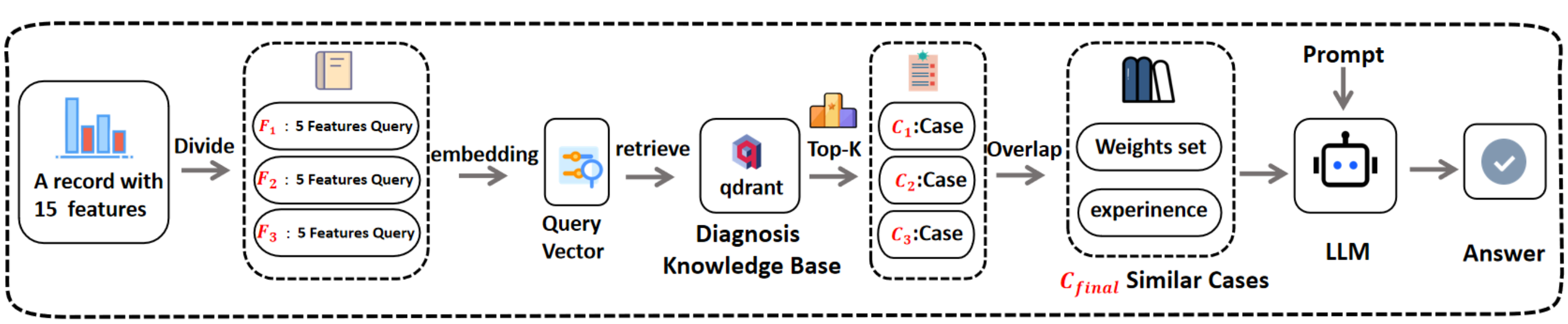

**Figure 3. The architecture for Prediction and Reasoning.** This diagram illustrates Module (III ). For a new patient with features, the Feature-Grouped Multi-Round Retrieval (FGMR) method queries the DKB to find a set of semantically similar cases (C final). The features of the new patient, along with the retrieved similar cases (including their feature weights and diagnostic text), are fed as a comprehensive prompt into the LLM. The LLM then performs a joint analysis to generate a final, context-aware prediction and a detailed report.

#### 2.5.1 Feature-Grouped Multi-Round Retrieval for Intersection Case Selection

We propose a method named Feature-Grouped Multi-Round Retrieval for Intersection Case Selection (FGMR) retrieval, to identify cases consistently detected across three feature subsets. Fifteen features are evenly divided into three groups ($F_1, F_2, F_3$). For each group will be sent to Qdrant as feature query to retrieve cases $C_K$, using BGE-Embedding-1024 and BEG-Reranker (Top-k, k=8) with a cosine similarity threshold 0.7. The final selection is the intersection of the three retrieval results:

$$C_{final} = C_1 \cap C_2 \cap C_3$$

ensuring cases are detected by all feature queries.

#### 2.5.2 LLM-Driven Sarcopenia Prediction and Report Generation

After applying the FGMR retrieval method and obtaining candidate similar cases that represent the correct diagnostic state, the large language model (LLM) conducts an internal evaluation to determine which cases are most relevant for further analysis. It then performs a feature-by-feature examination to derive the final weight for each clinical feature. These optimized weights are applied to compute the patient's sarcopenia probability.

By quantifying the contribution probabilities of feature interactions, the model identifies the relative influence of each feature and elucidates the specific pathways through which these features affect the diagnostic process. This dual-learning mechanism—leveraging probabilistic contributions to model feature importance and weight assignments to characterize feature interactions—enables the LLM to generate a structured diagnostic report that integrates SHAP-derived statistical evidence, reinforcement-trained reasoning, and case-based correlations.The pseudocode provided in Appendix A.5 Algorithm 2 and Algorithm 3.

The Sarcopenia Diagnosis Prediction Sample can be checked in Appendix A.4.

## IV. Result

### 3.1 Comparison of Prediction Results of Sarcopenia

Due to the extremely complex characteristics of sarcopenia and the existence of many data types, we use XGBoost as the teacher model of the large model. XGBoost supports multiple data types and allows the combination of tree models and linear models (referred to as GBDT+linear) within one model, which enhances the expressive ability of the model, further improves the prediction performance, and can flexibly handle complex nonlinear and linear problems.

In the prediction of sarcopenia using LLMs, for each case, multiple independent predictions are generated, and the single prediction achieving the highest performance—evaluated at its optimal threshold—is selected as the representative outcome. This approach ensures that the final prediction reflects the model's best achievable performance under the most favorable decision boundary. The results demonstrated consistent improvements under this strategy, as summarized in Table 1 and Figure 4 illustrates the prediction flow.

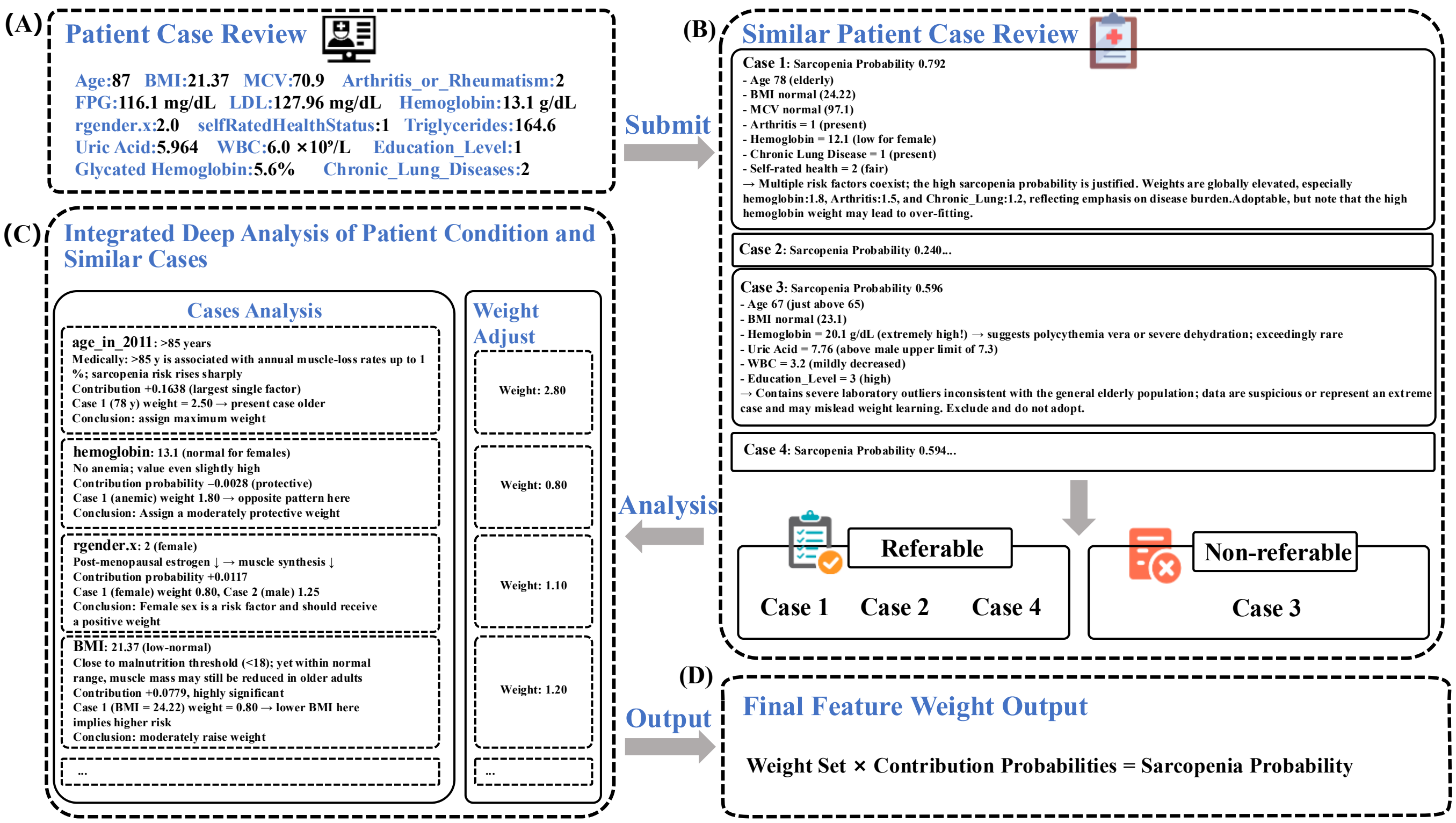

**Figure 4. The process of sarcopenia prediction in a specific case, consisting of four core parts.** (A) Patient Case Review: Presents the patient's clinical information such as age, BMI, biochemical indicators (e.g., fasting plasma glucose, low - density lipoprotein, etc.) and chronic diseases; (B) Similar Patient Case Review and Screening: After submitting the case in (A), similar cases are retrieved. Combined with the rationality judgment of cases, referable cases (Case 1, Case 2, Case 4) and non - referable cases (Case 3) are screened out; (C) Integrated Deep Analysis of Patient Condition and Similar Cases: Conducts case analysis on key features (age, hemoglobin, gender, BMI, etc.) and adjusts feature weights based on the analysis results; (D) Final Feature Weight Output: Calculates the sarcopenia probability through the logic of "Weight Set × Contribution Probabilities", fully presenting the whole process from case input, similar case matching and analysis to sarcopenia probability calculation

**Table 1.** Performance comparison of various models for sarcopenia prediction

| Model | Result Evaluation(normal/abnormal) (%) N=10 | | | | |
|---|---|---|---|---|---|
| | Method | Precision | Recall | F1score | Acc |
| XGBoost | / | 79.4/65.4 | 75.0/70.8 | 77.1/68.0 | 73.3 |
| Deepseek-R1-8B | / | 55.3/34.3 | 46.4/47.1 | 50.5/39.7 | 45.0 |
| Qwen3-plus | ACPB | 75.4±1.01/ 83.0±7.18 | 91.9±4.23/ 55.0±4.30 | 82.8 ± 1.30/ 65.8±1.46 | 77.2 ±1.11 |
| Deepseek-R1-8B | DKB | **77.8**±2.51/ 84.5±7.70 | 91.9±5.47/ **60.4**±7.15 | 84.2±2.18/ **69.9**±4.24 | **79.3** ±2.51 |
| Qwen3-plus | DKB | 77.4±**1.80**/ **85.2**±5.41 | **92.8±3.75**/ 59.2±**5.12** | **84.3±1.26**/ 69.5±**2.45** | **79.3** **±1.41** |

The table evaluates various models based on precision, recall, F1-score, and accuracy for both normal and abnormal classes. The proposed model, Deepseek-R1-8B and Qwen3-plus with reinforcement training, achieves the highest accuracy (79.3) and demonstrates superior performance across most metrics compared to the baseline XGBoost, LLMs without specialized training, LLMs only using XGBoost SHAP (ACPB).

### 3.2 Consistency Analysis

**Table 2.** Summary of consistency metrics for model output probabilities (N = 10, LLM= Qwen3-plus).

| Average Std | Average CV | ICC | Kappa Coefficient | Average Sample Consistency Rate |
|---|---|---|---|---|
| 0.230 | 38.4% | 0.956 | 0.842 | 0.955 |

To quantify prediction consistency under fixed input conditions, we evaluated both continuous probability outputs and binary classification outcomes across repeated model runs using an identical prompt. The standard deviation (SD) of predicted probabilities was 0.230 and the coefficient of variation (CV) was 38.4%, indicating relatively low dispersion and stable probabilistic behavior. The intraclass correlation coefficient (ICC) was 0.956, which—based on the classification criteria in [25]—reflects excellent reliability for continuous scores.

For binary classification at the predefined threshold, the Cohen's kappa coefficient was 0.842, denoting substantial agreement across runs [26]. Additionally, the mean sample-level consistency rate reached 0.9550, indicating that over 95.5% of individual cases were assigned identical classification labels in repeated executions. This high consistency underscores the model's reproducibility and robustness under controlled conditions, which is essential for clinical deployment and auditability.

To further support the consistency analysis, Table 3 summarizes the classification stability of the LLM-based model

**Table 3.** Summary of consistency metrics for model classification (N = 10, LLM=Qwen3-plus).

| Average AUC | Average Precision | Average Recall | Average F1 | Best Threshold |
|---|---|---|---|---|
| 0.760 ±0.010 | 0.852 ±0.054 | 0.592 ±0.051 | 0.695 ±0.025 | 0.697 ±0.037 |

To evaluate classification consistency under strictly controlled conditions, performance robustness was assessed by repeatedly executing the model using an identical prompt across multiple runs. The average area under the receiver operating characteristic curve (AUC) was $0.760 \pm 0.010$, indicating stable discriminative capability despite stochastic variations inherent to the model. Complementary metrics further confirmed this consistency: average precision was $0.852 \pm 0.054$, recall reached $0.592 \pm 0.051$, and the F1-score was $0.695 \pm 0.0252$. These results demonstrate balanced sensitivity and specificity across repeated trials under fixed input conditions. The optimal classification threshold, derived from prompt-consistent performance, was $0.697 \pm 0.037$, suggesting a reliable and reproducible decision boundary.

This prompt-invariant evaluation isolates model-internal randomness, allowing for a more precise assessment of classification stability. Such consistency is critical for clinical deployment, where reproducibility under fixed conditions ensures trustworthiness and auditability.

Under the optimal threshold, both the false positive rate (FPR) and true positive rate (TPR) exhibited noticeable fluctuations across prompt variations. This indicates that while the overall discrimination capacity—as reflected by stable AUC values—was preserved, the operating characteristics at the decision boundary remained prompt-sensitive. Such divergence suggests that the model does not merely replicate XGBoost outputs, but rather internalizes its attribution logic and dynamically adapts it during inference.

**Table 4.** Variations in Classification Operating Characteristics at the Optimal Threshold Across Different Prompts

| Prompt | Conservative | Normal | Aggressive |
|---|---|---|---|
| FPR | 0.056 | 0.139 | 0.306 |
| TRP | 0.583 | 0.708 | 0.708 |

Relative to the conservative setting, the aggressive prompt markedly increased the TPR while concurrently elevating the FPR, suggesting that prompt framing can modulate the model's decision threshold and risk propensity without materially affecting the AUC.

We assessed feature weight consistency across retrieved similar cases to evaluate the stability of attributions emerging from the latent space reconstruction process. For each query case, weight consistency was quantified using normalized metrics, including the coefficient of variation (CV), standard deviation, Cronbach's alpha, and cosine similarity. As shown in Appendix A.6 Figure 3, the majority of case groups reconstructed in latent space exhibited high internal consistency, with cosine similarity approaching 1.0 and Cronbach's alpha > 0.7, reflecting that the latent space effectively aligns feature importance patterns across similar cases. A smaller subset of case groups (e.g., Case IDs 42, 20, 37, 38, 46) displayed greater variability (CV > 0.15), suggesting residual heterogeneity in the latent space representation or sensitivity to the retrieval threshold. Overall, these results indicate that the observed stability and reproducibility of feature weights are direct consequences of the latent space–based fusion, demonstrating that the latent space reconstruction effectively governs the alignment of attribution patterns across similar-case clusters.

### 3.3 Contribution Probability using Sigmoid Analysis

To evaluate potential statistical bias in the proposed contribution probability calculation method, we conducted a comparative analysis between the predicted probabilities generated by the XGBoost model and the inferred probabilities derived from the training dataset. The inferred probabilities were defined as the sum of the contribution probability aligned with the ACPB and a fixed base probability of 0.5.

For each training sample, we computed the absolute difference between the two probability sets and summarized the distribution of these differences using descriptive statistics. The results are shown in Table 5.

Table 5. Statistical comparison of XGBoost output probabilities and CACS-inferred probabilities

| Bias Analysis | Mean | Std | Median | Min Bais | Max Bais |
|---|---|---|---|---|---|
| | 0.017 | 0.012 | 0.014 | 1.4e-4 | 0.063 |

The low mean absolute error (0.017) and narrow dispersion range indicate a high degree of agreement between the two probability sets, suggesting that the proposed contribution probability calculation method introduces negligible statistical bias and is both stable and reliable.

Notably, when we intentionally swapped the gender encoding, the large language model was able to detect that the probability contributions were inconsistent with established medical knowledge and accordingly made appropriate adjustments. Details of this process can be found in the Appendix A.7. This phenomenon demonstrates the model's capability to deeply integrate SHAP-based probability contributions with domain-specific medical knowledge through its own reasoning ability.

## V. Discussion

### Dual Preservation of Interpretability and Decision Correctness

Unlike conventional API-based integrations that transmit only final predictions, ACPB retains the verifiable logic of feature contributions, embedding interpretability as an intrinsic, operational element of the reasoning process rather than a post-hoc supplement and delivers two key benefits:

Semantic–Statistical Alignment – By creating the uniqueness of SHAP-derived feature-level contribution probabilities, ACPB ensures that the LLM's semantic reasoning space remains firmly anchored to the statistical inference space of the underlying TML model. This alignment constrains generative variability and mitigates the risk of unsupported reasoning paths.

Decision Fidelity – As evidenced in Table 5, only 0.017 absolute error, ACPB reproduces XGBoost outputs with negligible absolute deviation, quantitatively confirming that the statistical integrity of the original model is maintained throughout the transformation process.

Those two benefits make LLMs have great consistency in sarcopenia prediction compared with other diagnose LLMs, which can be checked in Table 3 and Table 4 that AUC stayed stable and risk propensity changed without materially affecting the AUC.

### From Redundant SHAP Recomputations to a Persistent, Extensible, and Auditable Feature-Level Attribution Repository

Traditional SHAP-based explanation workflows require re-running the TML model for each new instance, which is computationally expensive and produces only coarse-grained feature contribution insights that often lack the precision and contextual relevance needed for effective integration into LLM reasoning. In contrast, ACPB persistently stores fine-grained, feature-level SHAP contributions as structured, queryable evidence, eliminating redundant inference while providing reusable, context-aware attribution signals. By encoding these attributions in a model-agnostic, feature-centric format, ACPB supports incremental updates as models evolve or domains shift, enables seamless reuse across architectures and systems without retraining the original TML model, and allows attribution units to be cross-validated against external datasets or knowledge bases to meet audit and compliance requirements. Functioning as a persistent attribution library, ACPB maintains interpretability throughout the model lifecycle, thereby enhancing auditability, reproducibility, and trust in high-stakes decision-making contexts.

### Atomic Facts as Dual Anchors for Interpretability and Stability in ACPB

We define atomic facts as structured evidence units that decompose case-level attribution logic into precise, feature-level contributions.

Atomic facts serve as independent "proof points" for relevance assessment, ensuring that the model can determine, with precision, how each specific feature influences the outcome. By grounding similarity judgments in such traceable evidence, ACPB not only resolves the ambiguity inherent in conventional case-level matching but also establishes a transparent, auditable foundation for explaining feature–outcome relationships through the direct citation of atomic facts.

Our consistency analysis supports this claim: under repeated runs with identical prompts, the model maintained high reproducibility (ICC = 0.956, κ = 0.842, sample-level consistency = 95.5%), alongside stable discriminative capacity (AUC = 0.760 ± 0.010) [27]. While prompt framing modulated operating characteristics such as false positive and true positive rates, the decision boundary itself remained anchored to attribution evidence. This stabilizing effect parallels recent findings by Vladika [28], who showed that atomic fact checking in retrieval-augmented LLMs improves both explainability and reliability. In the context of ACPB, atomic facts thus serve a dual role: they provide transparent, auditable explanations, and they act as structural invariants that anchor LLM reasoning, enhancing robustness and trustworthiness in clinical deployment.

### Bridging Statistical Attribution and Semantic Reasoning for Interpretable Decision Pathways

Our work addresses two persistent challenges in explainable AI for high-stakes domains: the computational inefficiency and coarse granularity of traditional SHAP-based interpretability, and the limited numerical grounding of existing LLM–RL integration paradigms. By introducing ACPB, we replace per-sample SHAP recomputation with persistent, feature-level aggregate probability contributions, intentionally trading a small degree of individual-level precision for stable, reusable attribution signals. This "information degradation" is a deliberate structural choice that reduces noise from sample variability, supports robust cross-task alignment, and frees representational capacity for LLMs to inject domain knowledge, contextual adjustments, and adaptive weighting during reasoning. Building on this foundation, our CANDLE framework situates the LLM as the Actor in an actor–critic loop, where it consumes ACPB-derived attributions and clinical priors to propose feature-weight adjustments accompanied by natural language rationales. These adjustments are treated as concrete, optimizable actions, enabling fine-grained feedback at the feature level and aligning semantic reasoning with statistical evidence. We observe high consistency with cosine similarity approaching 1.0 and Cronbach's alpha > 0.7 in adjustment patterns across matched cases in Appendix A.6 Figure 3, indicating that the LLM generalizes attribution reasoning rather than producing arbitrary changes. This integration of quantitative grounding and semantic interpretability yields outputs that are both auditable and clinically meaningful, bridging the gap between opaque numerical models and transparent, feature-level decision pathways. In doing so, ACPB and CANDLE together form a persistent, verifiable interpretability layer that not only enhances model auditability and reproducibility but also transforms LLM reasoning into a structured, optimizable policy for decision support.

## Dual-Anchored Reward Function

In CANDLE's deployment, the dual-anchor design combines ACPB as a statistical anchor with foundational medical knowledge as a domain anchor to balance quantitative fidelity and semantic interpretability. The statistical anchor provides a stable, empirically validated baseline for feature-level attribution, constraining the decision space and ensuring consistency across cases. The domain anchor contextualizes these attributions, validating unusual patterns and enriching them with clinically meaningful narratives. Together, they enable rapid detection of abnormal features, guide targeted weight adjustments, and maintain both statistical rigor and clinical relevance throughout optimization. This integration not only accelerates convergence but also produces outputs that are auditable, interpretable, and actionable in high-stakes settings, positioning CANDLE as both a robust statistical detector and a transparent reasoning assistant.

### Limitations and Future Work

Despite the promising results, several limitations should be acknowledged. First, the current evaluation was conducted exclusively in the context of sarcopenia diagnosis. While this provides an initial proof of feasibility, extending the framework to a broader range of clinical scenarios is necessary to assess its generalizability across diverse disease domains. Second, CANDLE currently relies primarily on machine learning–derived attribution information, with only limited incorporation of clinical knowledge to enhance interpretability and semantic richness. Conflicts between clinical knowledge and attribution information can still arise, and the present design resolves such cases mainly by prioritizing attribution signals. Future work should therefore explore strategies to more organically integrate statistical attributions with clinical knowledge, thereby retaining the advantages of data-driven reasoning while ensuring clinical coherence. Third, the experiments employed XGBoost as the sole predictive model and SHAP as the attribution method. In principle, the framework can be validated with a wider variety of machine learning models—including deep neural networks—and alternative interpretability approaches such as LIME or Integrated Gradients, to assess robustness across modeling paradigms. Fourth, the current results remain at the proof-of-concept stage with limited validation data. Large-scale, multi-center studies using real-world clinical data will be required to rigorously evaluate robustness, cross-institutional generalizability, and clinical adoptability. Such studies will also help determine the framework's performance in heterogeneous patient populations under practical constraints. Fifth, in the consistency analysis, the number of repeated runs per sample was relatively not enough (n = 10) due to computational resource limitations. While this repetition count was sufficient to illustrate stability trends, a larger number of repetitions would provide narrower confidence intervals and more reliable statistical inference and We also observe that all models have optimal thresholds above 0.5 and above XGBoost optimal threshold—potentially reflecting design choices that favor admitting higher-probability predictions in the RL loop. Moreover, although the impact of different prompt styles on model performance was examined, systematic evaluation of prompt rephrasing and variability in clinical description formulations was beyond the scope of this study. Given prior evidence that such variations can influence large language model outputs—particularly near decision boundaries—future work should incorporate controlled prompt-rephrasing experiments and simulate realistic documentation variability to comprehensively assess robustness under diverse input conditions. Finally, while CANDLE demonstrates the feasibility of coupling XGBoost-derived attributions with large language models, the present study did not further investigate how LLMs concretely integrate their own reasoning capabilities with XGBoost outputs to realize measurable gains in classification accuracy. The analysis was also limited to samples correctly predicted by XGBoost, without examining misclassified cases, and it did not assess whether LLMs inherit or amplify biases present in the base model. Future work should address several key directions to strengthen and extend CANDLE. First, the framework should be evaluated across a broader range of clinical scenarios and disease domains to assess generalizability. Second, strategies for more seamless integration of statistical attributions with clinical knowledge should be developed, enhancing both interpretability and semantic richness while resolving potential conflicts between data-driven and domain-driven signals. Third, the approach should be validated with a wider variety of predictive models and interpretability methods, including deep neural networks, LIME, and Integrated Gradients, to ensure robustness across modeling paradigms. Fourth, large-scale, multi-center studies using real-world clinical data are needed to evaluate performance in heterogeneous populations and under practical constraints. Fifth, consistency analyses should be expanded with more repeated runs per sample and systematic evaluation of prompt variability to comprehensively assess stability and robustness. Finally, future work should investigate how LLMs can effectively integrate both correct and misclassified XGBoost outputs, elucidating the mechanisms by which attribution signals are internalized and assessing whether LLMs inherit or amplify base-model biases, thereby translating the hybrid design into measurable improvements in predictive accuracy and clinical utility.

## VI. Conclusion

This study introduces CANDLE that systematically embeds SHAP-derived attribution logic from traditional machine learning into LLM reasoning. Applied to sarcopenia diagnosis, CANDLE demonstrated superior accuracy, high reproducibility, and stable discrimination while preserving interpretability. By coupling statistical evidence with reinforcement-trained semantic reasoning, the framework effectively mitigates the long-standing trade-off between performance and transparency in clinical decision support. Beyond sarcopenia, CANDLE provides a scalable pathway for transforming machine learning outputs into structured, interpretable knowledge assets, offering a foundation for reliable, reproducible, and clinically aligned AI systems across diverse medical domains [29].

## Conflict of interest

The authors report no conflict of interest. We did not participate in recruiting the participants, as this article is based on publicly available data from the NHANES in the US. Since written informed consent was obtained from all participants in NHANES, this study was exempt from ethics review approval.

## Funding

This work is supported by the Ministry of Science and Technology of the People's Republic of China (STI2030-Major Projects2021ZD0201900), and the National Natural Science Foundation of China under Grants 12090052 and U24A2014.

## Data availability

Data described in the manuscript, code book, and analytic code will be made publicly and freely available without restriction at https://www.cdc.gov/nchs/nhanes/index.htm. The analytic code will be made available upon request pending application and approval from the corresponding author.

# Appendix

## Contents

# A Appendix

## A.0 Data preparation

The dataset for this study was derived from the China Health and Retirement Longitudinal Study (CHARLS), a nationwide survey examining health and retirement trends among middle-aged and elderly individuals in China. The analysis targeted participants free of sarcopenia at the baseline in 2011, with the incidence of sarcopenia tracked over a four-year period ending in 2015. Key variables selected for the study included demographic characteristics (age, gender), clinical measurements (height, weight, handgrip strength, skeletal muscle index [SMI]), laboratory test results (AST, ALT, creatinine, hemoglobin, albumin), and treatment-related factors (history of chemotherapy and surgical interventions). Missing data were handled using the K-Nearest Neighbors (KNN) imputation algorithm, optimized to maintain accuracy and data integrity. The resulting dataset was validated and processed using R (version 4.3.2), ensuring reproducibility and transparency throughout the preparation phase.

## A.1 Details of MODULE A Machine Learning Knowledge Extraction

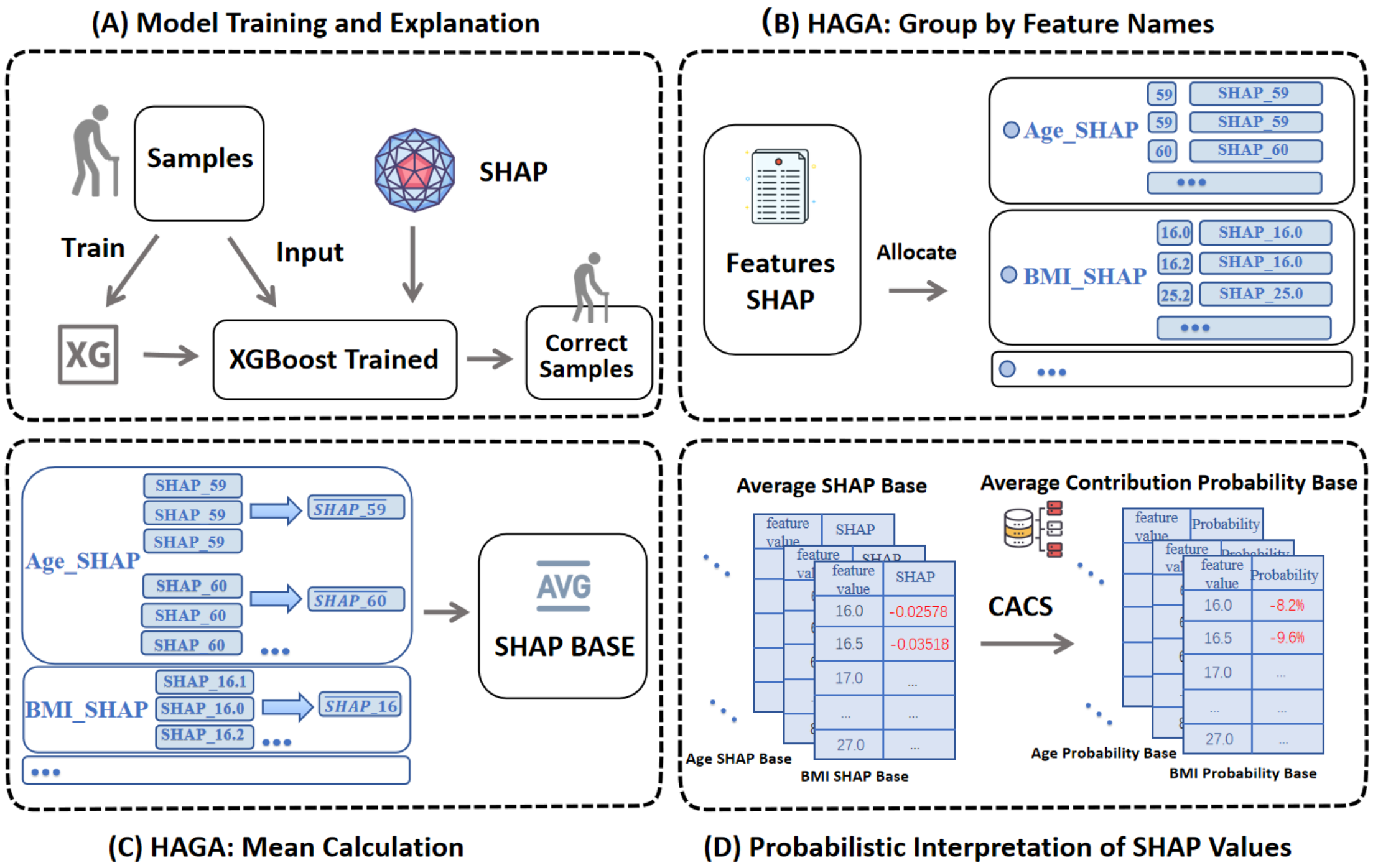

**Figure 1. workflow of the knowledge transformation pipeline.** This figure illustrates the four key steps in converting raw sample data into a probabilistic knowledge base for the LLM. It begins with (1) Data processing and Model Training using XGBoost and SHAP. This is followed by (2) HAGA: Group by Feature Names, where SHAP values are allocated to their respective features. Then, (3) HAGA: Mean Calculation computes the average SHAP value for discretized feature intervals. Finally, (4) Probabilistic Interpretation of SHAP Values uses the Contrastive Attribution via Sigmoid (CACS) method to transform these averaged SHAP values into contribution probabilities.

### *XGBoost training and SHAP Explanation*

In our methodology, we employ XGBoost-Regressor as the teacher model in our knowledge distillation framework, shown in Figure 2, Model Training, leveraging its demonstrated stability and predictive accuracy for sarcopenia probability estimation and use SHAP module to explain the prediction mechanism. The dataset undergoes stratified partitioning into three distinct subsets: a training set containing 360 samples (60% of total data), a validation set with 180 samples (30%), and a test set comprising 60 samples (10%). Through hyperparameter optimization with 190 boosting rounds and a maximum tree depth of 8, we developed a sarcopenia prediction model that achieves 90% training accuracy while maintaining 73% validation accuracy and 70% test accuracy. Then, we use SHapley Additive exPlanations (SHAP) to explain features by calculating SHAP values of every feature.

Following the development of our XGBoost-Regressor model and SHAP Explanation, we integrated them with a new approach to construct an interpretable knowledge base.

This systematic approach, consisting of Half-step Aligned Group Averaging and Probabilistic

Transformation, enables quantification of feature contributions and latent pattern extraction, thereby constructing an actionable knowledge base for LLM to assessment through evidence-based sarcopenia.

## *Half-step Aligned Group Averaging(HAGA)*

After former sections, we got a 324×15 Feature-SHAP Matrix:

$$\text{Feature} - \text{SHAP Matrix} = \begin{bmatrix} s_1 \\ s_2 \\ \vdots \\ s_{324} \end{bmatrix} = \begin{bmatrix} f_{1,1} & f_{1,2} & \cdots & f_{1,15} \\ f_{2,1} & f_{2,2} & & f_{2,15} \\ & \vdots & \ddots & \vdots \\ f_{324,1} & f_{324,2} & \cdots & f_{324,15} \end{bmatrix}$$

where $s_i$ is the i-th correct sample and $f_{i,j}$ is the j-th feature value with its SHAP value in the i-th correct sample.

And it is hard to directly utilize the SHAP values as a cross-model bridge due to the discrepancy between population-level statistics and individual feature distributions. To align SHAP values with both population-level statistics and individual feature distributions, we propose a three-stage discretization framework based on adaptive interval grouping, named Half-step Aligned Group Averaging, shown in Figure 1.

The core of this approach is to group feature values by feature names and compute mean SHAP values per group using predefined non-overlapping intervals with a fixed step size unit. The detailed workflow is outlined below:

## *HAGA Stage 1: Feature Grouping by Intervals*

**Group by Feature Names:**

Initially, for a given 324×15 Feature-SHAP Matrix, we carried out a transposition operation on it, thereby acquiring a Feature Set containing 15 sub-feature sets. Within each of these feature sets, there are 324 feature values accompanied by their respective corresponding SHAP values:

$$Feature - SHAP\ Matrix^T = Feature\ Set = \begin{bmatrix} F_1 \\ F_2 \\ \vdots \\ F_{15} \end{bmatrix} = \begin{bmatrix} f_{1,1} & f_{2,1} & \cdots & f_{324,1} \\ f_{1,2} & f_{2,2} & & f_{324,2} \\ & \vdots & \ddots & \vdots \\ f_{1,15} & f_{2,15} & \cdots & f_{324,15} \end{bmatrix}$$

Where $F_i$ is the $i$-th feature set and $f_{i,j}$ is the $j$-th feature value with its SHAP value in the $i$-th correct sample. Then, we can group feature and SHAP values in one certain feature set.

**Interval Construction:**

There two types of features, continuous features and integer features for interval construction, this study proposes a unified interval partitioning rule to achieve discretization and structured representation of feature values.

For continuous features, an equidistant midpoint partitioning approach is adopted: a midpoint sequence $X = \{x_m | x_m = 0.5m, m \in Z\}$ is generated with a step size of 0.5. Each midpoint defines a closed interval of width 0.5, specifically $[x_m - 0.25,\ x_m + 0.25]$, which encompasses all continuous values within ±0.25 of $x_m$. For example, midpoint 0.5 corresponds to the interval [0.25, 0.75], midpoint 1.0 corresponds to [0.75, 1.25], and so on.

For integer features, the interval partitioning simplifies to an exact matching rule: the feature value must strictly equal the specified midpoint $x_m$ (where $x_m \in Z$). That is, only when the feature value $V = x_m$ is the sample assigned to the interval group centered at $x_m$. This rule ensures that the discretization results of integer features have clear boundaries and uniqueness.

**Interval Match:**

To establish a bijective correspondence between feature values and discrete intervals, the following formal grouping rule is defined: For any feature value $V$, if it satisfies one of the following conditions, it is assigned to the corresponding interval group $g$ :

Continuous feature value: There exists a midpoint $x_m$ such that $x_m - 0.25 \leq V \leq x_m + 0.25$;

Integer feature value: $V = x_m$ (where $x_m$ is an integer midpoint).

This mapping rule ensures that each feature value uniquely belongs to one interval group, and each interval group contains all feature values and their corresponding SHAP values that meet the condition. Through this operation, the original feature space is transformed into a set of finite discrete intervals, providing structured input for subsequent knowledge distillation.

After interval partitioning and grouping mapping, Feature Set is organized into the following matrix form:

$$Assigned\ Feature\ Set = \begin{bmatrix} F_1 \\ F_2 \\ \vdots \\ F_{15} \end{bmatrix} = \begin{bmatrix} g_{1,1} & g_{1,2} & \cdots & g_{1,k_1} \\ g_{2,1} & g_{2,2} & & g_{2,k_2} \\ & \vdots & \ddots & \vdots \\ g_{15,1} & g_{15,2} & \cdots & g_{15,k_{15}} \end{bmatrix}$$

Where $F_i$ is the $i$-th feature interval group set, $g$ is the interval group in one certain interval and $k_i$ is the length of interval in the $i$-th feature interval group set.

Subsequently, based on the discretized feature intervals, we obtained the corresponding sets of SHAP and feature values for each interval. These sets reflect the degree of influence that feature values within specific ranges have on sarcopenia prediction, providing preliminary insights into local feature effects.

However, when attempting to estimate the SHAP value of a new feature value by referencing similar feature values within existing interval groups, we still encounter the challenge of ambiguous multi-feature-values matching in one certain interval group —where the new feature value could be plausibly closed to many feature values.

## *Stage 2: HAGA: Interval Mean Aggregation*

**Interval Mean Aggregation**

We calculate the arithmetic mean of all values $V$ within the same interval to serve as the representative value for that group to solve the ambiguous multi-feature-values matching challenge.

By establishing clear - cut interval boundaries and discretization rules, this method effectively eliminates the ambiguity associated with traditional rounding techniques. Simultaneously, it ensures computational efficiency and maintain interpretability. It finds wide applications in data standardization, noise filtering, and feature engineering tasks.

For all SHAP values within the $i$-th interval compute the mean:

$$\bar{s_i} = \frac{1}{N} \sum_{v \in Group_i} V$$

where $N$ is the count number of SHAP values in the $i$-th group.

Through former two stages, we systematically processed the SHAP values to enhance useability of SHAP knowledge. The methodology yields 15 specialized feature SHAP knowledge sub-bases:

$$SHAP\ Knowledge\ Base = \begin{bmatrix} b_1 \\ b_2 \\ \vdots \\ b_{15} \end{bmatrix} = \begin{bmatrix} \bar{s}_{1,1} & \bar{s}_{1,2} & \cdots & \bar{s}_{1,k_1} \\ \bar{s}_{2,1} & \bar{s}_{2,2} & & \bar{s}_{2,k_2} \\ \vdots & & \ddots & \vdots \\ \bar{s}_{15,1} & \bar{s}_{15,2} & \cdots & \bar{s}_{15,k_{15}} \end{bmatrix}$$

Where $b_i$ is the $i$-th feature SHAP knowledge sub-bases, $\bar{s}$ is the mean SHAP value in one certain interval and $k_i$ is the length of mean SHAP value in the $i$-th feature SHAP knowledge sub-bases. Each feature SHAP knowledge sub-base contains: (1) feature values discretized with fixed step sizes, and (2) their corresponding averaged SHAP values within defined ranges.

This grouping mechanism enables effective alignment between individual feature instances and their subgroup SHAP distributions, facilitating coherent information utilization. However. it is still difficult for LLMs to utilize SHAP values.

**Probabilistic Transformation for ACPB**

To address the interpretability of SHAP values, we chose to transform them into a structured format optimized for LLMs, rather than relying on resource-intensive domain knowledge integration. This approach minimizes token consumption while enabling the LLM to generate contextually relevant explanations from preprocessed feature importance signals.

We implement a probabilistic transformation framework, combining Sigmoid Normalization:

$$\sigma(\varphi_i) = \frac{1}{1 + e^{-\varphi_i}}$$

where $\varphi_i$ represents the original SHAP value, converting feature contributions into normalized probabilities (0-1 range)

For a model $f$ and instance $x$, SHAP value $\varphi_i$ satisfy:

$$f(x) = \varphi_0 + \sum_{i=1}^{M} \varphi_i(x)$$

where $\varphi_0 = E[f(x)]$ is the base value.

For specific feature impact analysis, we compute differential contributions through Contrastive Attribution Calculation using Sigmoid (CACS):

Let $z = \varphi_0$ The probability contribution $p_j$ of $j$ interval mean SHAP value is defined as:

$$p_i = \sigma(z + \varphi_j) - \sigma(z)$$

Where $\sigma$ is the sigmoid function.

By computing probability contributions for all interval mean SHAP values, we establish an Average Contribution Probability Base for LLMs:

$$CACS(SHAP\ Knowledge\ Base) = \text{Average Contribution Probability } Base$$

Specifically:

$$Average\ Contribution\ Probability\ Base = \begin{bmatrix} b_1 \\ b_2 \\ \vdots \\ b_{15} \end{bmatrix} = \begin{bmatrix} p_{1,1} & p_{1,2} & \cdots & p_{1,k_1} \\ p_{2,1} & p_{2,2} & & p_{2,k_2} \\ & \vdots & \ddots & \vdots \\ p_{15,1} & p_{15,2} & \cdots & p_{15,k_{15}} \end{bmatrix}$$

Where $b_i$ is the $i$-th feature sub-bases in Average Contribution Probability Base, $p$ is the contribution probability and $k_i$ is the length of contribution probability values in the $i$-th feature sub-bases in Average Contribution Probability Base.

This framework enables systematic characterization of feature impacts for new incoming feature values, providing a probabilistic foundation for interpretable model analysis.

## A.2 Key Components of the Actor-Critic Learning Loop

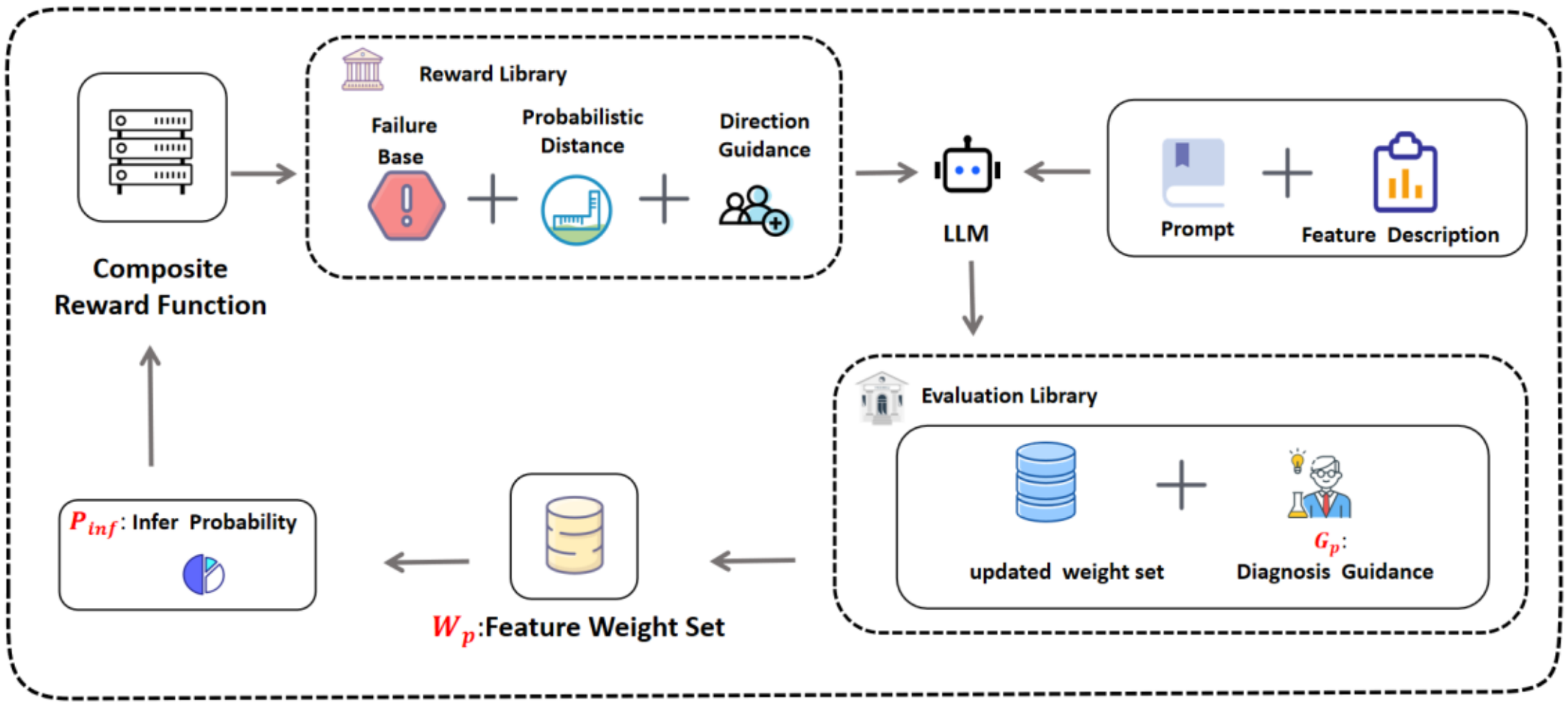

**Figure 2. Detailed workflow of Actor-Critic Learning Loop.** If their discrepancy exceeds 5%, the Critic generates a Reward Library to guide the LLM (as the Actor), which in turn produces an Evaluation Library with updated feature weights and diagnostic guidance. The loop iterates, generating new Infer Probabilities, until the discrepancy falls below 5%—at which point the refined diagnostic text and metadata are stored.

## *Actor (LLM)*

The LLM serves as the Actor, with core functions including:

**Infer Probability Optimization**: Based on the Reward Library (feedback from Critic), feature descriptions, and clinical knowledge, adjust feature weights to refine Infer Probability.

**Diagnostic Guidance Generation**: Synthesize the Evaluation Library, which includes updated feature weights (reflecting adjusted feature importance) and diagnostic guidance (explaining the reasoning behind feature contributions).

**Latent Space Restoration & Augment**: By integrating XGBoost's feature contribution probabilities (from ACPB) with its own contextual reasoning, the LLM compensates for the degradation of latent space caused by ACPB, enhancing the representation of patient-specific features.

## *Critic (Composite Reward Function)*

The Composite Reward Function acts as the Critic, with two pivotal roles:

**Deviation Quantification**: Calculate the difference ( $diff$ ) between Infer Probability and True Probability, where $diff = \frac{|True\ Probability - Infer\ Probability|}{True\ Probability}$. This quantifies the relative deviation to determine whether the reinforcement learning loop needs to be activated (triggered when $diff > 5\%$).

**Reward Library Generation**: Generate a structured Reward Library to guide the LLM, which includes:

**Failure Case Replay**: Stores recent cases with $diff$ > 5% (up to 3 cases, using FIFO replacement) to provide historical error references.

**Probability Distance Metrics**: Quantified diff values to reflect the magnitude of deviation.

**Direction Guidance Signals**: Indicator of whether the deviation is in the correct direction relative to the decision boundary (0.5), helping the LLM adjust optimization trajectories.

## *Workflow of Actor-Critic Loop*

The workflow of Module (B) is detailed steps as follows:

**Input and Initial Feature Processing**: For a given patient record, extract 15 structured clinical features. Match these features with ACPB to extract corresponding contribution probabilities, constructing the Patient Feature Table ($T_p$) (the cornerstone of subsequent probability inference).

Initial Infer Probability Calculation: Use an initial weight set (all weights = 1, ensuring initial inference is solely based on XGBoost's prior knowledge without LLM bias) to calculate the initial Infer Probability ($P_{inf}$) via weighted summation of $T_p$'s contribution probabilities.

Deviation Judgment: The Critic (Composite Reward Function) calculates the deviation ($diff$) between initial $P_{inf}$ and XGBoost's True Probability ($P_{xgb}$):

If $diff$ ≤ 5%: The initial inference is sufficiently accurate. Directly save $T_p$ (as retrieval vector), initial weights, and diagnostic guidance (generated by LLM based on $T_p$ and feature descriptions) as metadata.

If diff > 5%: Activate the Actor-Critic loop for iterative optimization.

**Loop Iteration:**

a. The Critic generates a Reward Library (including Failure Case Replay, Probability Distance Metrics, Direction Guidance Signals) based on the current diff.

b. The Actor (LLM) synthesizes an Evaluation Library using the Reward Library and feature descriptions, outputting updated feature weights and diagnostic guidance.

c. Update the weight set with the new weights, recalculate $P_{inf}$ using $T_p$ and the updated weights.

d. The Critic recalculates diff, if diff ≤ 5%, save $T_p$, updated weights, and diagnostic guidance, then terminate the loop. Otherwise, repeat steps a-d.

Output Storage: High-quality diagnostic texts (integrating XGBoost's logic and LLM's reasoning) and associated metadata (weights, guidance) are stored for subsequent retrieval and application.

### ***Core Value: Latent Space Restoration***

Information Degradation by ACPB: For a given clinical feature $f$, consider a subgroup $G \subseteq Dataset$ containing n patient samples $\{x_1, x_2, \dots, x_n\}$. Each sample $x_i \in G$ has an original contribution probability $c_{f,i}$ for feature $f$, reflecting its individual-specific impact on diagnosis.

The ACPB constructs the representative contribution probability for feature f in subgroup $G$ via intra-group averaging: $\bar{c}_f = \frac{1}{n}\sum_{i=1}^{n} c_{f,i}$ where $\bar{c}_f$ denotes the averaged contribution probability stored in ACPB.

This averaging operation inherently causes information degradation, quantified by the **loss of individual-specific variance**: The variance of original contribution probabilities within subgroup $G$ is: $\sigma_f^2 = \frac{1}{n}\sum_{i=1}^{n}(c_{f,i} - \bar{c}_f)^2$. Here, $\sigma_f^2$ measures the dispersion of individual contributions around the group average. When $\sigma_f^2 > 0$, it indicates the presence of sample-specific information. However, this variance is discarded in ACPB ($\sigma_f^2 = 0$), as only $\bar{c}_f$ is retained, leading to information degradation.

This degradation blurs the latent space's representation of individual patient characteristics, necessitating restoration via the Actor-Critic loop, as LLMs reasoning compensates for the lost variance $\sigma_f^2$.

## A.3 Prompt

### ***Training: weight adjustment***

You, as a highly professional doctor with decades of professional experience, adjust the weights of features based on various types of information. The feature weights adjusted by you will be used to calculate the correct probability of sarcopenia.

Your Task: Please conduct in-depth thinking, step by step. You will obtain information similar to that of the patient, but not identical to the patient's. You need to utilize your medical knowledge, historical weight information, and the information provided to you to update the patient's feature weights.

Your Output: A set of feature weights. The feature weights must be positive numbers, cannot be 0, and must not exceed 3 at maximum. The formula for sarcopenia probability is: Sarcopenia Probability = Sum (Feature Weight * Feature Contribution Probability).

Other Explanations:

Feature Contribution Probability: The degree to which a feature contributes to the patient's risk of developing sarcopenia under its current feature value. A positive value indicates that the feature will increase the probability of sarcopenia, while a negative value indicates that the feature will decrease the probability of sarcopenia.

Feature Weight: Should be greater than or equal to 0, cannot be negative, and must be a positive number. Increasing the weight of a feature with a positive contribution degree will increase the patient's probability of sarcopenia; increasing the weight of a feature with a negative contribution degree will decrease the patient's probability of sarcopenia.

Sarcopenia Probability: A numerical value ranging from 0 to 1.

The feature weight [weight] in the sarcopenia calculation formula is not the standard case contribution degree.

Score: Feedback from a reinforcement learning expert, which can quantify the gap between your results and the standard answers. You should conduct precise optimization based on the optimization directions provided in the reinforcement learning expert's feedback.

Important Methods:

1. Increasing the weight of a feature with a positive contribution probability will increase the sarcopenia probability; decreasing the weight of a feature with a positive contribution probability will decrease the sarcopenia probability.
2. Increasing the weight of a feature with a negative contribution probability will decrease the sarcopenia probability; decreasing the weight of a feature with a negative contribution probability will increase the sarcopenia probability.

Follow the steps below:
Step 1 Receive Data: As a doctor, you will receive a comprehensive patient feature analysis table: {table}.
Step 2 Receive Historical Versions: From a doctor's perspective, you will receive the [{{sarcopenia inference information}}] of the {{previous three versions}} related to sarcopenia and the [{{adjustment information}}] from the reinforcement learning expert:
{{
Third-to-last version (oldest version): {last_three}\n

Second-to-last version (middle version): {last_two}\n

Last version (latest version): {last}]\n
}}

Patient Feature Description: {des}
Step 3 Analyze Historical Versions to Derive New Weights: As a doctor, analyze and understand the [{{adjustment information}}, {{feature weights}}, and {{scores}}] of the three versions from a doctor's perspective, adjust the feature weights, and derive new feature weights.

Principles for Adjusting Feature Weights ###: {{[

Principle 0: You should be guided by the sarcopenia probability, so that the adjusted feature weights can correctly calculate the sarcopenia probability. Sarcopenia Probability = Sum (Feature Weight * Feature Contribution Probability).
Principle 1: You can only adjust and modify the {{feature weights (feature weights must be positive numbers, cannot be 0, and must not exceed 3 at maximum)}}, and try not to make the weights of too many features too large. You are not allowed to adjust or modify the {{standard case contribution degree (impact)}} in any way.
Principle 2: Maintain the distribution of feature weights. Do not make the weights of most features greater than 1, nor make the weights of most features less than 1. Important note: A normal distribution of feature weights should have some weights greater than 1 and some less than 1.
Principle 3: Your understanding of the impact of patient features on the sarcopenia probability is derived from the positive or negative nature of the feature contribution probability. For example, if a patient is 62 years old but the feature contribution probability (of age) is negative, it means that being 62 years old will decrease the sarcopenia probability. Therefore, increasing the feature weight of age will decrease the sarcopenia probability.
}}

Comprehensive Patient Feature Analysis Table:
{table}
{weights}
{input}

Output Format: x.xx is only an example; you need to fill in the actual numbers for the weights. Feature weights must be positive numbers, cannot be 0, and must not exceed 3 at maximum:
weights = {{
'age_in_2011.x': x.xx,
'BMI': x.xx,
'MCV': x.xx,
'Arthritis_or_Rheumatism': x.xx,
'FPG': x.xx,
'LDL': x.xx,
'hemoglobin': x.xx,
'selfRatedHealthStatus': x.xx,
'Triglycerides': x.xx,
'uric_acid': x.xx,
'WBC': x.xx,
'Education_Level': x.xx,
'glycated_hemoglobin': x.xx,

'rgender.x': x.xx,
'Chronic_Lung_Diseases': x.xx
}}

## Prediction

You are a highly accomplished medical expert with profound expertise in the field of sarcopenia diagnosis. Relying on your extensive experience and exquisite medical knowledge, you can determine the feature weight for each feature based on the detailed medical records of the patient and relevant information of similar cases.

Below are the detailed procedures and specific requirements you need to follow during the diagnosis process:

**Diagnosis Procedures**

**1. Review of Patient's Condition**

**Explanation of Similar Cases Content**:

**Patient's Medical Information**: Specific information of similar patients (A positive contribution probability indicates that the patient is more likely to develop sarcopenia; a negative contribution probability indicates that the patient is more likely to be in good health).

**Corresponding Feature Weight**: The weight corresponding to each feature, which serves to amplify or reduce the feature's contribution probability.

Patient's Condition: {table}

Feature Description: {desc}

**2. Analysis of Similar Cases (please conduct the most thorough analysis possible)**

Similar Cases: {record}

Principles:

Principle 1: As a professional doctor, you must analyze the patient's sarcopenia condition using your professional medical knowledge.

Principle 2: The contribution probability of a patient's feature can reveal the effect of that feature on the probability of sarcopenia. A positive feature contribution probability will increase the risk of sarcopenia, while a negative feature contribution probability will reduce the risk of sarcopenia. You must combine your medical knowledge with the numerical value of the feature contribution degree.

Principle 3: Please note that "feature contribution probability" and "feature weight" are two distinct concepts. Feature weight refers to the degree to which the feature's contribution probability is amplified or reduced.

Principle 4: Please exclude any cases you consider unreasonable from the similar cases and refrain from adopting them.

Principle 5 (Most Important Principle): During the analysis, you must combine the specific feature contribution probability with the weights of similar cases, and clearly demonstrate the process of referencing each feature.

**3. Joint Analysis of Patient's Condition and Similar Cases**

Principles:

Principle 1: The analysis is not merely a match between similar cases and the patient's condition, but rather a medical analysis that utilizes medical knowledge and patient data.

Principle 2: The condition of similar cases is only similar to the patient's condition to a certain extent; there is no need for direct matching.

Please select the feature weight you deem correct for each feature.

Strictly follow the format below for outputting feature weights:
{{
'age_in_2011.x': x.xx,
'BMI': x.xx,
'MCV': x.xx,
'Arthritis_or_Rheumatism': x.xx,
'FPG': x.xx,
'LDL': x.xx,
'hemoglobin': x.xx,
'selfRatedHealthStatus': x.xx,
'Triglycerides': x.xx,
'uric_acid': x.xx,
'WBC': x.xx,
'Education_Level': x.xx,
'glycated_hemoglobin': x.xx,
'rgender.x': x.xx,

'Chronic_Lung_Diseases': x.xx
}}

## A.4 Sarcopenia Diagnosis Prediction Sample

**Step 1: Patient Condition Review and Medical Interpretation**

The patient's basic information is as follows:

- **Age**: 87 years old (age_in_2011.x = 87.0)
  → Belongs to the advanced-age group. Individuals over 68 years old are at high risk of sarcopenia, and the risk increases exponentially with age. The **contribution probability is +0.1638**, which significantly positively drives the risk of sarcopenia, consistent with pathophysiological mechanisms.
- **BMI = 21.37**
  → At the lower limit of the "normal" range (20-24), close to the underweight threshold (<18), suggesting potential insufficient nutritional reserve or low muscle mass. The **contribution probability is +0.0779**, supporting its role as a risk factor.
- **MCV = 70.9**
  → Significantly lower than the normal range (80–100 fL), indicating **microcytic anemia**, which may result from iron deficiency, chronic diseases, or malnutrition—all of which are associated with impaired muscle synthesis. However, the standard contribution probability is **-0.0171** (negative), seemingly "protective," which contradicts medical common sense. Data model bias or confounding effects should be suspected, and clinical judgment is required for further evaluation.
- **Arthritis_or_Rheumatism = 2** (Healthy)
  → No arthritis/rheumatism, indicating a protective state. The contribution probability is **-0.0162**, which is reasonable.
- **FPG = 116.1 mg/dL**
  → Exceeds the normal range (<99), indicating **impaired fasting glucose (IFG)**. Insulin resistance can lead to increased muscle protein breakdown. The contribution probability is **+0.0155**, with a correct direction.
- **LDL = 127.96 mg/dL**
  → Higher than the ideal value (<100), suggesting atherosclerotic risk, which indirectly affects muscle perfusion and mitochondrial function. The contribution probability is **+0.0005**, small but with a reasonable direction.
- **Hemoglobin = 13.1 g/dL**
  → Slightly low for males (normal: 13.8–17.2 g/dL), close to the anemia threshold; normal for females. Patient's gender: **Female (rgender.x=2)? Note the coding instructions!**

Patient's rgender.x = 2.0 → **Female**

Therefore, a hemoglobin level of 13.1 g/dL is normal to high for females (normal: 12.1–15.1 g/dL) and should not be considered anemia. The contribution probability is **-0.0028** (negative), i.e., "reducing sarcopenia risk," which is reasonable.

- **selfRatedHealthStatus = 1**
  → Self-rated as "unhealthy," strongly indicating decreased physical function, reduced activity, and heavy psychological burden—an independent predictor of sarcopenia. The contribution probability is **+0.0468**, which is highly reasonable.
- **Triglycerides = 164.6**
  → Mildly elevated (>150 is abnormal), reflecting metabolic disorders, often accompanied by insulin resistance. The contribution probability is **+0.0179**, which is reasonable.
- **Uric Acid = 5.964**
  → Within the normal range for females (1.6–6.0), at the upper limit, close to hyperuricemia. Uric acid has a certain antioxidant effect, but excessively high levels suggest metabolic syndrome. The contribution probability is **+0.0001**, weakly positive, which is acceptable.
- **WBC = 6.0 × $10^9$/L**
  → Within the normal range (3.5–9.5 × $10^9$/L for females), with no obvious inflammation. The contribution probability is **+0.0094**, showing a slight pro-inflammatory tendency, which may reflect subclinical inflammation.
- **Education_Level = 1**
  → Low education level, usually associated with low health literacy, unbalanced diet, and weak awareness of exercise. The contribution probability is **+0.0071**, which is reasonable.
- **Glycated Hemoglobin = 5.6%**
  → Normal (<5.7%), but close to the pre-diabetic threshold. The contribution probability is **+0.0058**, weakly positive, possibly due to proximity to the threshold.
- **Chronic_Lung_Diseases = 2**
  → No chronic lung disease, indicating a protective factor. The contribution probability is **-0.0020**, which is reasonable.

**Step 2: Similar Case Analysis (Denoising and Screening)**

We evaluated the rationality of four similar cases and decided whether to adopt their weights as references.

✅ **Case 1: Sarcopenia Probability 0.792**

- Age: 78 years old (advanced age)
- Normal BMI (24.22)
- Normal MCV (97.1)
- **Arthritis = 1 (Affected)**
- **Hemoglobin = 12.1 g/dL (Low for females)**
- **Chronic Lung Disease = 1 (Affected)**
- Self-rated health = 2 (Fair)

→ Multiple risk factors coexist, and the high sarcopenia probability is reasonable. The overall weights are relatively high, especially for hemoglobin: 1.8, Arthritis: 1.5, and Chronic_Lung: 1.2, reflecting emphasis on disease burden. **Adoptable, but attention should be paid to potential overfitting due to the excessively high hemoglobin weight.**

✅ **Case 2: Sarcopenia Probability 0.240**

- Age: 60 years old (<65, low risk)
- BMI: 26.95 (Overweight)
- High LDL (137.6)
- High hemoglobin (16.7)
- Low education level (1)
- No chronic diseases, self-rated health: good (3)

→ Young age + good health status, so low risk is reasonable. The weight for Education_Level: 1.6 is relatively high, but overall reasonable. **Adoptable.**

❌ **Case 3: Sarcopenia Probability 0.596**

- Age: 67 years old (just over 65)
- Normal BMI (23.1)
- **Hemoglobin = 20.1 g/dL (Extremely elevated!)** → Obvious abnormality, possibly polycythemia vera or severe dehydration, extremely rare
- Uric Acid = 7.76 (exceeds the upper limit of 7.3 for males)
- WBC = 3.2 (mildly decreased)
- Education_Level = 3 (High)

→ There are severe laboratory abnormalities, which are inconsistent with the characteristics of the general elderly population. The **data is suspicious or an extreme case**, which may mislead weight learning. **Excluded and not adopted.**

✅ **Case 4: Sarcopenia Probability 0.594**

- Age: 67 years old
- BMI: 31.1 (Obese)
- Extremely high LDL (164.3)
- Other indicators basically normal
- No chronic diseases, self-rated health = 3

→ Although obesity may "mask" muscle loss (i.e., "sarcopenic obesity"), the age has just reached the threshold and there are no other obvious risk factors. A sarcopenia probability of approximately 0.59 is slightly high but still acceptable. The weights are relatively balanced. **Can be used as a reference.**

**Step 3: Combined Analysis of Patient Condition and Similar Cases (Medical-Driven + Data-Assisted)**

We will assign weights to each feature based on the following principles:

1. **Weights should reflect the pathological importance of the feature in the context of this patient**
2. **Combine the direction and magnitude of the standard contribution probability**
3. **Refer to the weight trends of reasonable cases and eliminate the interference of extreme values**
4. **Prioritize biological rationality over pure numerical matching**

◆ **age_in_2011.x: 87 years old (Extremely advanced age)**

- Medically: Individuals over 85 years old are at extremely high risk of sarcopenia, with a muscle loss rate of up to 1–2% per year
- Contribution probability: +0.1638 (the largest single item)
- Among similar cases: Case 1 (78 years old) has a weight of 2.50 → This patient is older, so the weight should be higher
- **Conclusion**: Assign the highest weight

  ➤ **Weight: 2.80**

◆ **BMI: 21.37 (Normal to low)**

- Close to the malnutrition threshold (<18); although not reaching it, muscle loss can still occur in the elderly with a "normal BMI"
- Contribution probability: +0.0779, significant
- Case 1 (BMI=24.22) has a weight of 0.80 → This patient has a lower BMI and higher risk

- **Conclusion**: Appropriate increase

  ➤ **Weight: 1.20**

◆ **MCV: 70.9 (Significantly decreased)**

- Microcytic anemia → Suggests iron deficiency or anemia of chronic disease → Inhibits muscle synthesis
- However, the contribution probability is **-0.0171** (negative), conflicting with medical logic
- Possible model confusion: Some healthy individuals have low MCV without anemia, or data bias
- Combined with medical knowledge, it **should be a positive risk factor**
- Case 1 (normal MCV) has a weight of 0.70; Case 3 (excluded) has a weight of 2.10 (unreasonable)
- **Conclusion**: Correct model bias and assign a moderate positive weight

  ➤ **Weight: 1.30**

◆ **Arthritis_or_Rheumatism: 2 (Healthy)**

- No joint disease → Protective
- Contribution probability: -0.0162
- Case 1 (affected) has a weight of 1.50 → By reverse deduction, healthy individuals should have reduced weight impact
- **Conclusion**: Assign a low weight to reflect "absence of risk"

  ➤ **Weight: 0.60**

◆ **FPG: 116.1 (IFG)**

- Abnormal glucose metabolism → Insulin resistance → Increased muscle protein breakdown
- Contribution probability: +0.0155
- Case 2 has a weight of 1.15, Case 4 has a weight of 1.00
- **Conclusion**: Moderate weight

  ➤ **Weight: 1.10**

◆ **LDL: 127.96 (Elevated)**

- Cardiovascular risk → Affects microcirculation → Decreased muscle oxygen supply
- Contribution probability is minimal (+0.0005) but with a reasonable direction
- Both Case 2 and Case 4 have a weight of 1.40
- **Conclusion**: Maintain neutral to high

➤ **Weight: 1.10**

◆ **hemoglobin: 13.1 (Normal for females)**

- No actual anemia, even slightly high
- Contribution probability: -0.0028 (protective)
- Case 1 (anemic) has a weight of 1.80 → Opposite for this patient
- **Conclusion**: Assign a moderate protective weight

  ➤ **Weight: 0.80**

◆ **selfRatedHealthStatus: 1 (Self-rated as unhealthy)**

- Strong signal of psychological and functional decline
- Contribution probability: +0.0468 (the second largest)
- Case 1 (score 2) has a weight of 0.70 → This patient has a worse score, so the weight should be higher
- **Conclusion**: Significant increase

  ➤ **Weight: 1.60**

◆ **Triglycerides: 164.6 (Elevated)**

- Marker of metabolic syndrome
- Contribution probability: +0.0179 (relatively significant)
- Case 2 and Case 3 have weights of 0.95 and 0.75 → Relatively low
- **Conclusion**: Increase to reflect metabolic risk

  ➤ **Weight: 1.20**

◆ **uric_acid: 5.964 (Close to the upper limit for females)**

- Dual effects: Antioxidation vs. metabolic disorder
- Contribution probability is minimal (+0.0001)
- Case 3 (excluded) has a weight of 2.15 → Unreliable
- **Conclusion**: Slightly positive

  ➤ **Weight: 1.00**

◆ **WBC: 6.0 (Normal)**

- No obvious inflammation
- Contribution probability: +0.0094
- Case 1 and Case 2 have weights of 0.80 and 0.90

- **Conclusion**: Neutral

  ➤ **Weight: 0.90**

◆ **Education_Level: 1 (Low)**

- Poor health behaviors → Decreased diet quality, exercise, and compliance
- Contribution probability: +0.0071
- Case 2 has a weight of 1.60 (young but low education still indicates high risk) → Supports its importance
- **Conclusion**: Increase

  ➤ **Weight: 1.30**

◆ **glycated_hemoglobin: 5.6%**

- Normal but close to the threshold
- Contribution probability: +0.0058
- Case 3 (excluded) has a weight of 1.50 → Too high
- **Conclusion**: Moderate

  ➤ **Weight: 1.00**

◆ **rgender.x: 2 (Female)**

- Decreased estrogen in postmenopausal females → Decreased muscle synthesis
- Contribution probability: +0.0117
- Case 1 (female) has a weight of 0.80, Case 2 (male) has a weight of 1.25
- **Conclusion**: Females are a risk factor, so assign a positive weight

  ➤ **Weight: 1.10**

◆ **Chronic_Lung_Diseases: 2 (Healthy)**

- No lung disease → Protective
- Contribution probability: -0.0020
- Case 1 (affected) has a weight of 1.20 → Opposite for this patient
- **Conclusion**: Assign a low weight

  ➤ **Weight: 0.60**

✅ **Final Feature Weight Output (Strict Format)**

json

{

"age_in_2011.x": 2.80,

"BMI": 1.20,

"MCV": 1.30,

"Arthritis_or_Rheumatism": 0.60,

"FPG": 1.10,

"LDL": 1.10,

"hemoglobin": 0.80,

"selfRatedHealthStatus": 1.60,

"Triglycerides": 1.20,

"uric_acid": 1.00,

"WBC": 0.90,

"Education_Level": 1.30,

"glycated_hemoglobin": 1.00,

"rgender.x": 1.10,

"Chronic_Lung_Diseases": 0.60

}

📌 **Summary Explanation**

This weight system is based on:

- **Priority of medical logic**: Corrected potential reverse biases of variables such as MCV and hemoglobin in the model;
- **Individualized adjustment**: Strengthened weights for core risk factors including 87-year-old female, poor self-rated health, and metabolic abnormalities;

- **Denoising processing**: Excluded extreme Case 3 to avoid misleading;
- **Dynamic balance**: Achieved a balance between data contribution and clinical experience, ensuring the interpretability and robustness of the weights.

These weights can be used for subsequent optimization of risk scoring models or development of individualized intervention strategies.

## A.5 Pseudocode

Here are pseudocodes of three Algorithms.

Algorithm 1 Cross - Model Knowledge Distillation Framework

```
Require: A record with 15 features (f_1, f_2, ...,
f_15), weight set (w_1, w_2, ..., w_15)
       Average   Contribution   Probability   Base
(ACPB)
       Pre-trained LLM
       Reward threshold ε

Ensure: Saved high-quality diagnostic texts

1: for each patient record R in the dataset do
2:       (ConPro_1,  ConPro_2,  ...,  ConPro_15)  ← R
Nearest  Match  with  ACPB  3:  Patient  Feature  Table
T_p ← (ConPro_1, ConPro_2, ..., ConPro_15) /UR
4:     Initial Weight Set ← set all weights to 1
5:     Initial P_inf ← ProbCalculation(T_p, initial
weight set)
6:      diff ← Calculate diff ← Composite Reward
Function(initial P_inf, P_true)
7:     if diff > ε then
8:        Diagnosis Guidance G_p ← LLM ← Prompt(T_p,
Feature Description)
9:        Save(Vector: T_p, Metadata: G_p, W_p)
10:    else
11:       while diff ≥ ε do
12:             Reward Library ← Composite Reward
Function(initial P_inf, P_true)
```

```
13:                 Evaluation Library ← LLM ← Prompt(Reward Library, Feature Description)
14:            Updated Weights, Diagnosis Guidance G_p ← Evaluation Library
15:            Weight Set W_p ← Updated Weights
16:            P_inf ← ProbCalculation(T_p, Weight Set)
17:            diff ← Calculate diff(P_inf, P_true)
18:            if diff ≤ ε then
19:                Save(Vector: T_p, Metadata: G_p, W_p)
20:                Break
21:            end if
22:        end while
23:    end if
24: end for
```

Algorithm 2 Prediction & Context-Aware Reasoning Framework

```
1: Input: Feature vector X = [x_1, x_2, ..., x_15]
2: Output: Diagnosis Prediction via LLM
3: Divide X into 3 groups: F_1, F_2, F_3, where F_i ∈ R^5
4: for i = 1 to 3 do
5:     Encode feature group F_i into query vector q_i = Embed(F_i)
6:     Retrieve top-k similar cases: C_i = TopK(Retrieve(q_i))
7: end for
8: Similar Cases ← Compute intersection of retrieved cases: C_final = ∩^3_{i=1} C_i
9: Weight Set W_p, Diagnosis Guidance G_p ← Similar Cases
10: Diagnosis Prediction ← LLMs ← Prompt(W_p, G_p)
```

Algorithm 3 LLM-Driven Sarcopenia Prediction and Report Generation.

```
1: Input: Patient data D_p, FGMR retrieval method R
2: Output: Final probability P_sarc and diagnostic report R_diag
3: C ← R(D_p)                  ▷ Retrieve similar cases using FGMR
4: H_0 ← LLM_Assess(D_p)       ▷ Initial hypothesis by LLM
```

```
H ← Refine(H_0, C)                ▷ Refine hypothesis
using retrieved cases
W ← GenerateWeights(H, C)         ▷ Generate
personalized weights
P_sarc ← ComputeProbability(D_p, W)   ▷ Calculate
sarcopenia probability
F_c ← QuantifyContributions(D_p, W)   ▷ Feature
contribution probabilities
F_c ← ModelFeatureImportance(F_c, W)  ▷ Evaluate
feature influence and interactions
R_diag ← GenerateReport(H, F_c, C)    ▷ Generate
structured diagnostic report
return P_sarc, R_diag
```

## A.6 Result

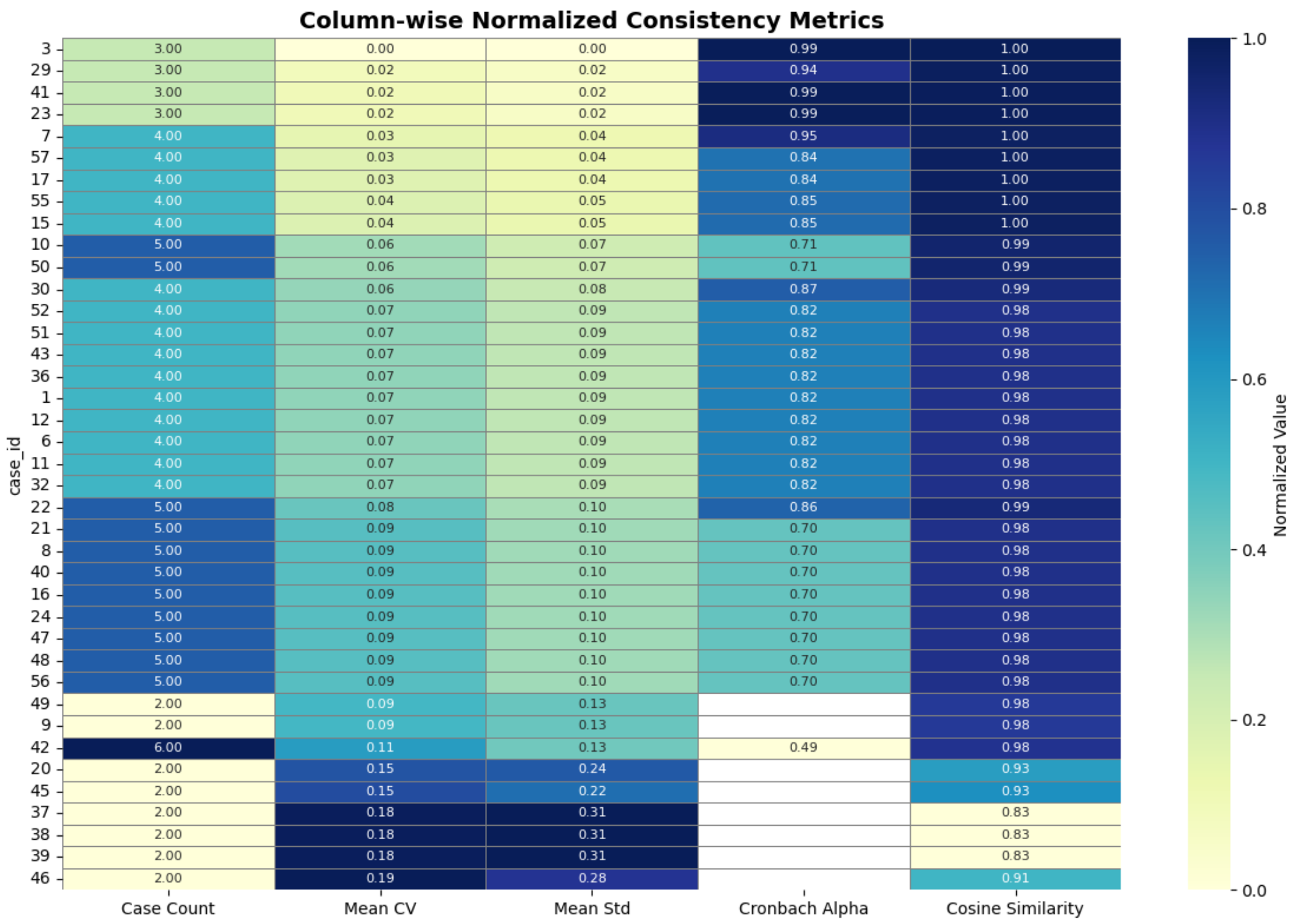

**Figure 3. Column-wise normalized consistency metrics for retrieved similar cases. Each row represents a query case and columns summarize the normalized metrics used to evaluate feature weight consistency: case count, mean coefficient of variation (CV), mean standard deviation (Std), Cronbach's alpha, and cosine similarity. Most case clusters exhibit high cosine similarity (≈1.0) and Cronbach's alpha (>0.7), suggesting robust weight alignment among retrieved similar cases.**

## A.7 Results of Gender Encoding Swap

| |
|---|
| Case1:<br><br>**14. rgender.x: 1 (Female)**<br><br>- Clinical perspective: Females (especially those aged 70 and above) have a higher risk of sarcopenia than males.<br><br>- However, the impact value is -0.0083 (incorrectly assumes that females are healthier).<br><br>- Similar cases: Case 1 (Female): 1.50, Case 2 (Female): 1.15 → Correctly reflect the high risk in females.<br><br>- **Model gender bias must be corrected.**<br><br>- **Conclusion: High weight → 1.40** |
| Case2:<br><br>**rgender.x = 1 (Female)**<br><br>- Clinically: A sharp decline in estrogen in postmenopausal females leads to reduced muscle synthesis, altered fat distribution, and decreased bone mineral density → accelerating the progression of sarcopenia.<br><br>- Epidemiology shows that the incidence rate of sarcopenia in females is higher than that in males at advanced age.<br><br>- **impact: -0.00834** (negative contribution) → ❌ Wrong direction!<br><br>> Model misguidance occurs again: Being female should increase the risk rather than decrease it.<br><br>- Must be corrected based on medical knowledge → **Increase the weight to reflect the actual risk**. |
| Case3:<br><br>14. `rgender.x`: 1 (Female)<br><br>- Postmenopausal estrogen decline → reduced muscle synthesis and increased fat replacement<br><br>- Incidence of sarcopenia in women over 65 is higher than in men<br><br>- impact=-0.0083 (negative contribution) → **Model error!**<br><br>- Medically, being female is a risk factor, not a protective one<br><br>- In Case 1, males were assigned 1.05 → females should have a higher weight<br><br>- **Model bias must be corrected** |

- **Reasonable weight: 1.15**

Case4:

14. `rgender.x`: **1.05**

- Value = 1 (Female): Postmenopausal estrogen decline accelerates muscle loss

- Although the impact is -0.0083 (negative value), medically, the risk of sarcopenia in females (especially elderly females) is not lower than that in males

- Multiple studies have shown that females experience faster decline in muscle mass

- The model may underestimate the risk in females due to sample bias

- **The weight should be increased based on medical facts**

- **Set: 1.05**

## A.8 Explanation of Nouns

**ACPB (Average Contribution Probability Base):** A foundational library storing contribution probabilities of clinical features, used to match patient records and extract initial feature contribution probabilities.

**Infer Probability:** The probability inferred by the framework, calculated based on the Patient Feature Table and feature weight set. The probability calculation formula as follows:

$$Infer\ Probability = Base\ Probability + \sum_{i}^{15}(Contribution\ Probability_i \times Weight_i)$$

where $Base\ Probability$ equals 0.5.

**True Probability:** The reference probability output by XGBoost, serving as the benchmark for evaluating the accuracy of Infer Probability.

**Actor:** Refers to the pre-trained Large Language Model (LLM), responsible for generating diagnostic guidance, updating feature weights, and outputting the Evaluation Library to optimize Infer Probability.

**Critic:** Refers to the Composite Reward Function, responsible for quantifying the deviation between Infer Probability and True Probability, and generating the Reward Library to guide the Actor's optimization.

**Reward Library:** The output of the Critic, consisting of three components: Failure Case Replay (historical cases with large deviations), Probability Distance Metrics (quantified deviation values), and Direction Guidance Signals (optimization direction for LLM). Probability Distance Metrics and Direction Guidance formulas shown as below:

$$Probability\ Distance\ Metrics = \begin{cases} f(\cdot), & diff > 0.05\ and\ \ S(\cdot) > 0 \\ 10, & diff \leq 0.05\ and\ \ S(\cdot) > 0 \\ 0, & S(\cdot) < 0 \end{cases}$$

$$Direction\ Guidance = \begin{cases} Your\ inferred\ probability\ is \begin{cases} significantly\ higher\ than\ actual\ levels \\ significantly\ lower\ than\ actual\ levels \end{cases}, & diff > 0.05\ and\ \ S(\cdot) > 0 \\ \text{Prediction direction correct with acceptable deviation,} & diff < 0.05\ and\ \ S(\cdot) > 0 \\ \text{Warning: Prediction contradicts factual direction. Re} - \text{examine decision basis,} & S(\cdot) < 0 \end{cases}$$

where $diff = \frac{|True\ Probability - Infer\ Probability|}{True\ Probability}$ quantifies the relative deviation between predicted and true probabilities; $S(\cdot) = (True\ Probability - 0.5) \times (Infer\ Probability)$ determines whether the prediction aligns with the true decision boundary (0.5); $f(\cdot) = \frac{True\ Probability}{|True\ Probability - Infer\ Probability|}$ inverse-proportional penalties on high-deviation predictions to enhance error sensitivity.

**Evaluation Library:** The output of the Actor, containing updated feature weights and diagnostic guidance, used to iteratively adjust Infer Probability.

**Latent Space:** A hidden feature space integrating XGBoost's structured logic and LLM's reasoning capabilities. It degrades due to ACPB's inherent traits (designed to accommodate LLM reasoning) and is restored through the Actor-Critic loop.